\documentclass{article}

\PassOptionsToPackage{numbers, compress}{natbib}
\usepackage[preprint]{neurips_2026}

\usepackage[utf8]{inputenc} % allow utf-8 input
\usepackage[T1]{fontenc}    % use 8-bit T1 fonts
\usepackage{hyperref}       % hyperlinks
\usepackage{url}            % simple URL typesetting
\usepackage{booktabs}       % professional-quality tables
\usepackage{amsfonts}       % blackboard math symbols
\usepackage{nicefrac}       % compact symbols for 1/2, etc.
\usepackage{microtype}      % microtypography
\usepackage{xcolor}         % colors
\usepackage[table]{xcolor}
\usepackage{array}
\usepackage{times}
\usepackage{makecell}
\usepackage{graphicx}
\usepackage{float}
\definecolor{ORANGEII}{HTML}{F0A64D}

\newcolumntype{L}[1]{>{\raggedright\arraybackslash}m{#1}}
\newcolumntype{C}[1]{>{\centering\arraybackslash}m{#1}}

\usepackage{enumitem}
\usepackage{pifont}
\usepackage{amsmath} 
\usepackage{pdfpages}
\usepackage{multirow}
\usepackage{subfig}
\usepackage[capitalize]{cleveref}
\usepackage{xcolor}
\usepackage{tikz}
\usepackage{pdfrender}
\usepackage{xspace}
\usepackage{hyperref}

\definecolor{MethodBlue}{HTML}{153D7A}
\definecolor{MethodOrange}{HTML}{F0A64D}

\newsavebox{\gradienttextbox}

\newcommand{\method}{\textsc{Forcing-KV}\xspace}

\newcommand{\gray}[1]{\cellcolor{gray!10}\textcolor{gray}{#1}}

\newcommand{\orangecell}[1]{\cellcolor{ORANGEII!20}{#1}}

\usepackage{siunitx}
\usepackage{pifont}
\usepackage{titlesec}

\newcommand{\gainnote}{%
  \makebox[0pt][l]{\textcolor{gainblue}{\raisebox{0.15ex}{\scriptsize\ding{115}}}}%
}

\definecolor{gainblue}{HTML}{153D7A}
\definecolor{uclablue}{rgb}{0.15, 0.45, 0.68}

\hypersetup{
    colorlinks=true,
    urlcolor=MethodBlue
}

\hypersetup{
    breaklinks,
    citecolor=uclablue,
    colorlinks=true,
    linkcolor=uclablue
}

\usepackage{graphicx}
\usepackage{tikz}

\begin{document}

\AddToShipoutPictureFG*{%
  \AtPageUpperLeft{%
    \hspace{3.7cm}%
    \raisebox{-2.65cm}{%
      \includegraphics[height=1.2cm]{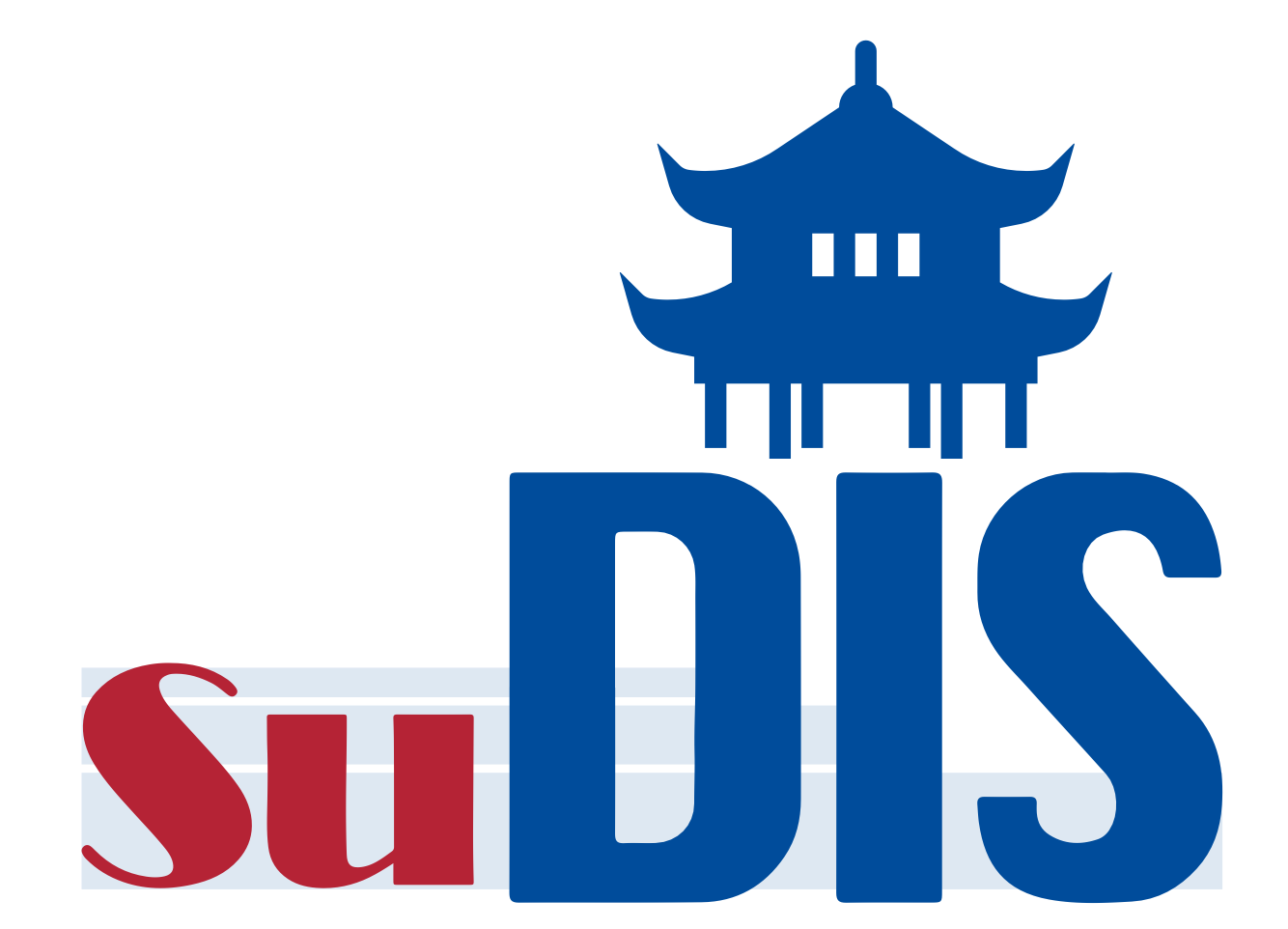}%
    }%
  }%
}

\AddToShipoutPictureFG*{%
  \AtPageUpperLeft{%
    \hspace{14cm}%
    \raisebox{-2.65cm}{%
      \includegraphics[height=0.6cm]{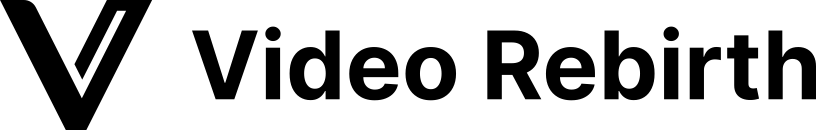}%
    }%
  }%
}

% \title{\texorpdfstring
%   {\method{}: Hybrid KV Cache Compression for Efficient Autoregressive Video Diffusion Models}
% }

\title{\sffamily\bfseries
  {Forcing-KV: Hybrid KV Cache Compression for Efficient Autoregressive Video Diffusion Models}
}

\author{
\begin{tabular}{c}
\textbf{Yicheng Ji}$^{1,2}$ \quad
\textbf{Zhizhou Zhong}$^{2,3}$ \quad
\textbf{Jun Zhang}$^{1}$ \quad
\textbf{Qin Yang}$^{2}$ \quad
\textbf{Xitai Jin}$^{2}$ \\[0.35em]
\textbf{Ying Qin}$^{4}$ \quad
\textbf{Wenhan Luo}$^{3}$ \quad
\textbf{Shuiyang Mao}$^{2}$ \quad
\textbf{Wei Liu}$^{2}$ \quad
\textbf{Huan Li}$^{1,\dagger}$ \\[0.8em]
{\large
$^{1}$\textbf{ZJU} \quad
$^{2}$\textbf{Video Rebirth} \quad
$^{3}$\textbf{HKUST} \quad
$^{4}$\textbf{BJTU}
} \\[0.4em]
\multicolumn{1}{c}{\footnotesize{$\dagger$ Corresponding Author}}
\end{tabular}
}

\maketitle

\projectpage{\url{https://zju-jiyicheng.github.io/Forcing-KV-Page}}
\emailaddress{\url{jiyicheng.cs@zju.edu.cn}}

\begin{abstract}
Autoregressive (AR) video diffusion models adopt a streaming generation framework, enabling long-horizon video generation with real-time responsiveness, as exemplified by the \textit{Self Forcing} training paradigm. 
However, existing AR video diffusion models still suffer from significant attention complexity and severe memory overhead due to the redundant key-value (KV) caches across historical frames, which limits scalability.
In this paper, we tackle this challenge by introducing KV cache compression into autoregressive video diffusion. 
We observe that attention heads in mainstream AR diffusion models exhibit markedly distinct attention patterns and functional roles that remain stable across samples and denoising steps.
Building on our empirical study of head-wise functional specialization, we divide the attention heads into two categories: \textit{static heads}, which focus on transitions across autoregressive chunks and intra-frame fidelity, and \textit{dynamic heads}, which govern inter-frame motion and consistency.
We then propose \method, a hybrid KV cache compression strategy that performs structured static pruning for static heads and dynamic pruning based on segment-wise similarity for dynamic heads.
While maintaining output quality, our method achieves a generation speed of over \textbf{29} frames per second on a single NVIDIA H200 GPU along with \textbf{30\%} cache memory reduction, delivering up to \textbf{1.35$\times$} and \textbf{1.50$\times$} speedups on LongLive and Self Forcing at 480P resolution, and further scaling to \textbf{2.82$\times$} speedup at 1080P resolution.

\end{abstract}

\vspace{-2pt}

\begin{figure}[H]
    \centering
    \includegraphics[width=0.87\linewidth]{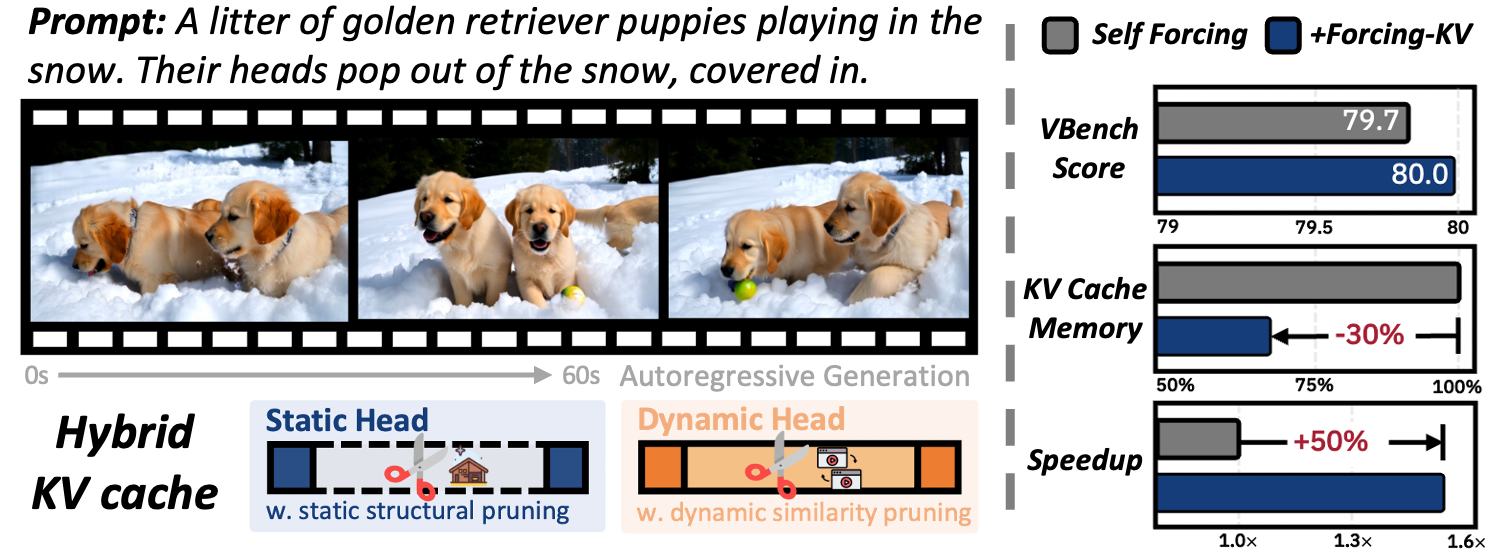}
    \caption{Overview of \method. We apply static structural pruning and dynamic similarity pruning to different heads, accelerating inference, reducing cache memory while improving quality.}
    \label{fig:teaser}
\end{figure}

\section{Introduction}
\label{sec:introduction}
% 双向模型
% Video diffusion models~\cite{Peebles2022DiT,lin2024opensoraplanopensourcelarge,kong2025hunyuanvideosystematicframeworklarge,wan2025wan} have greatly advanced text-to-video generation, enabling the production of realistic videos with high visual quality and rich temporal dynamics. However, their bidirectional attention mechanism denoises all frames simultaneously, constraining generation length and limiting interactive applications. 
% %引出自回归模型
% To address this limitation, recent studies distill pretrained bidirectional diffusion models into few-step autoregressive student models (AR diffusion models)~\cite{yin2024onestepdiffusiondistributionmatching,yin2025slow,teng2025magi,chen2025skyreels,huang2025selfforcing,yang2025longlive}, leveraging the \textit{Self Forcing}~\cite{huang2025selfforcing,cui2025self} training paradigm to mitigate error accumulation. By introducing key-value (KV) cache mechanisms, these models enable streaming, chunk-wise video generation, where each chunk attends to preceding chunks through sliding window attention.
Autoregressive (AR) video diffusion~\cite{huang2025selfforcing,yang2025longlive,yin2025slow,teng2025magi,chen2025skyreels,zhang2025framepack} has recently emerged as a compelling paradigm for efficient, streaming text-to-video generation. Unlike conventional bidirectional video diffusion models~\cite{Peebles2022DiT,lin2024opensoraplanopensourcelarge,kong2025hunyuanvideosystematicframeworklarge,wan2025wan} that denoise all frames simultaneously, AR video diffusion models produce video chunk by chunk, with each new chunk conditioned on previously generated video content via a key-value (KV) cache.
This paradigm enables long-horizon, variable-length video generation with interactive inputs, while reducing both attention complexity and the latency to the first generated content.
Mainstream approaches build upon the \textit{Self Forcing}~\cite{huang2025selfforcing,cui2025self} training paradigm, performing self-rollout during training to mitigate error accumulation, as exemplified by the broader family of ``forcing'' methods~\cite{huang2025selfforcing,yang2025longlive,cui2025self,lu2025reward,liu2025rolling,yi2025deep,yesiltepe2025infinity,xiao2025knot,huang2025live,li2026stable,cui2026lol} that have shown strong performance.

% 自回归模型的效率问题
However, existing mainstream AR video diffusion models still suffer from substantial attention complexity and severe memory overhead due to the heavy KV cache of historical chunks~\cite{yang2025longlive,lu2025reward,chen2026past}.
As video generation accumulates over time, the currently generated chunk is forced to attend to increasingly long and redundant visual context, which substantially reduces efficiency especially for long-horizon and high-resolution videos.
For instance, generating a 30-second video at 1080P resolution with Self Forcing~\cite{huang2025selfforcing} takes over 2 minutes on a single NVIDIA H200 GPU, corresponding to a generation speed of 1.71 FPS considering only the overhead within the diffusion transformer (DiT). Moreover, the KV cache alone consumes more than 60 GB of GPU memory in this setting, which poses a major obstacle to deployment in memory-constrained scenarios.

%现有加速方法
To achieve real-time inference, studies have explored sparse attention~\cite{lv2026light,agarwal2026monarchrt} and feature caching~\cite{maflow,samuel2026fast} techniques for AR video diffusion models.
Although effective, such methods neither reduce memory overhead nor operate on the KV cache, which is a distinctive structural component of AR video diffusion models. Recently, Dummy Forcing~\cite{guo2026efficient} observes that certain heads in AR diffusion models concentrate primarily on the currently generated chunk, and accordingly discards the historical context for those heads.
However, it lacks a detailed analysis of the functional heterogeneity across attention heads, and its aggressive compression results in degraded temporal dynamics and discontinuity across chunks (i.e., flickering and broken transitions at chunk boundaries), as shown in~\Cref{sec:experiment}.
We posit that for AR video diffusion models, effective context utilization is key to both quality and efficiency. This raises a pivotal question:

\textit{\textbf{Does autoregressive video diffusion model exhibit distinctive patterns in its KV cache utilization?}} 
    
Our findings suggest an affirmative answer. 
We observe markedly distinct attention patterns and functional roles across the attention heads of AR video diffusion models.
Through a series of careful empirical ablation studies in~\Cref{sec:observation}, we categorize the attention heads into the two categories.
\textit{Static heads} consistently attend to the current chunk and the most recent frame, which we denote as the \emph{transition anchor frame}, to preserve intra-frame fidelity and visual continuity across autoregressive chunks.
\textit{Dynamic heads} capture inter-frame correspondences across the same spatial regions, governing subject consistency and motion dynamics.
Moreover, we find that this head division remains stable across different samples and denoising steps, and generalizes broadly across multiple AR video diffusion models.
Based on these observations, we propose \method, a hybrid KV cache compression method for AR video diffusion models that decouples static structural patterns from dynamic context utilization.

\method first introduces a one-shot, model-level \textbf{offline head profiling} procedure (see~\Cref{sec:method_head_profile}) that identifies static and dynamic heads based on frame-wise attention mass.
Subsequently, we apply a hybrid KV cache compression strategy. For static heads, we adopt \textbf{static structural pruning} (see~\Cref{sec:method_static_compression}) to consistently preserve the transition anchor frame and prune distant frames. For dynamic heads, we employ \textbf{dynamic similarity pruning} (see~\Cref{sec:method_dynamic_compression}), which computes segment-wise similarity between adjacent frames in the KV cache to retain temporally evolving content while pruning redundant and unchanged content.
To summarize, our main contributions are:
\begin{itemize}[itemsep=0pt, topsep=0pt, leftmargin=19pt]
    \item[(1)] \textit{Novel Pattern Discovery}: We uncover a universal head specialization pattern shared by mainstream autoregressive video diffusion models: transitions across autoregressive chunks are mediated by static heads that concentrate on the transition anchor frame, whereas long-horizon consistency and dynamics are sustained by dynamic heads through inter-frame attention.

    \item[(2)] \textit{Hybrid KV cache Compression}: Building upon this, we propose \method, a compression strategy that preserves structurally critical content for static heads while applying dynamic similarity pruning for dynamic heads, decoupling static patterns from dynamic context utilization.
    % , achieving substantial compression while maintaining generation quality.

    \item[(3)] \textit{Extensive Experiments}: Evaluations across models, benchmarks, generation lengths, and resolutions show that \method is both high-fidelity and efficient: While maintaining quality, \method achieves up to \textbf{1.35$\times$} and \textbf{1.50$\times$} speedups along with \textbf{30\%} cache memory reduction on LongLive and Self Forcing at 480P resolution, further scaling to \textbf{2.82$\times$} at 1080P.
\end{itemize}

% \section{Related Work}
% \label{sec:related_work}
% We present core related work here and provide a detailed discussion in~\Cref{app:related_work}.
% %
% A growing number of works explore autoregressive diffusion modeling~\cite{zhang2025framepack,chen2025skyreels,teng2025magi,yin2025causvid} for fast long video generation.
% %
% Self Forcing~\cite{huang2025selfforcing} mitigates train-test
% discrepancy by performing self-rollout and LongLive~\cite{yang2025longlive} further extends this framework through KV recaching and long-horizon fine-tuning.
% Recent work~\cite{lu2025reward,liu2025rolling,cui2025self,yesiltepe2025infinity,xiao2025knot,huang2025live,li2026stable,cui2026lol,xiang2026pathwise,cai2026mode} has focused on generating minute-long videos, most of which build upon the \textit{Self Forcing} training paradigm. 
% These efforts reflect a broader trend toward long-horizon video generation where KV cache utilization is critical for scalability and efficiency.
% %
% To accelerate AR diffusion models, studies~\cite{gao2025ca2,agarwal2026monarchrt,lv2026light,samuel2026fast,maflow,xu2026sparseforcingnativetrainable,ranganath2026kvcachequantizationselfforcing} have explored feature caching and sparse attention but most do not alleviate cache size or memory overhead.
% Dummy Forcing~\cite{guo2026efficient} compresses the attention window for certain heads, but it lacks a detailed characterization of the attention patterns and functional roles of heads, which leads to discontinuities across chunks and drops in temporal dynamics.

\section{Related Work}
\label{app:related_work}

\paragraph{Video Diffusion Models.}
Video diffusion models have evolved from bidirectional, one-shot generation to autoregressive, streaming generation.
Early bidirectional video diffusion models~\cite{lin2024opensoraplanopensourcelarge,kong2025hunyuanvideosystematicframeworklarge,wan2025wan} are typically built upon the Diffusion Transformer (DiT)~\cite{Peebles2022DiT} architecture, enabling high-quality and controllable video generation.
To address the high cost of bidirectional denoising and support long-horizon video generation, a growing number of works turn to autoregressive diffusion modeling~\cite{zhang2025framepack,chen2025skyreels,teng2025magi}.
To further reduce denoising steps, CausVid~\cite{yin2025causvid} reformulates bidirectional diffusion into causal generation through distribution
matching distillation. Self Forcing~\cite{huang2025selfforcing} mitigates train-test
discrepancy by performing self-rollout during the training stage, and LongLive~\cite{yang2025longlive} further extends this framework through KV recaching and long-horizon fine-tuning. Krea-Realtime-14B~\cite{erwann2025krea} scales video generation to 14B parameters.
More recently, a growing body of work~\cite{lu2025reward,liu2025rolling,cui2025self,yesiltepe2025infinity,xiao2025knot,huang2025live,zhang2025framepack,li2026stable,cui2026lol,xiang2026pathwise,cai2026mode,meituanlongcatteam2025longcatvideotechnicalreport,yuan2026helios,cai2026mode} has focused on generating minute-long videos. Representative methods include Rolling Forcing~\cite{liu2025rolling}, Reward Forcing~\cite{lu2025reward}, Infinite-Rope~\cite{yesiltepe2025infinity}, and Self Forcing++~\cite{cui2025self}, most of which build upon the \textit{Self Forcing} training paradigm. 
% Some study~\cite{cai2026mode} attempt to decouple local fidelity from long-term coherence, which aligns with our observation of functional specialization among attention heads.
%
These efforts reflect a broader trend toward long-horizon video generation and the potential for a \textit{train-long–test-long} strategy, in which KV cache size and memory overhead are critical factors for scalability and efficiency.

\paragraph{Efficient Video Generation.}
Video diffusion models are computationally expensive due to heavy attention computation and multi-step denoising. For bidirectional models, inference is typically accelerated through sparse attention~\cite{xisparse,yang2025sparse,xu2026sparseforcingnativetrainable}, linear attention~\cite{chen2025sana}, quantization~\cite{zhang2024sageattention2}, and feature caching~\cite{liu2025timestep} techniques.
%
% For autoregressive models, specialized sparse attention~\cite{agarwal2026monarchrt,lv2026light,samuel2026fast} and feature caching~\cite{maflow} techniques have been explored.
Recently, several studies~\cite{gao2025ca2,agarwal2026monarchrt,lv2026light,samuel2026fast,maflow,xu2026sparseforcingnativetrainable,ranganath2026kvcachequantizationselfforcing} have attempted to tailor these acceleration techniques to the characteristics of AR video diffusion models.
% In autoregressive models, the context length varies across autoregressive steps, requiring attention kernels that can dynamically adapt. Moreover, mainstream autoregressive video diffusion models are few-step, which further limits the direct applicability of feature caching.
%
However, AR video diffusion models natively rely on \textit{KV cache} for streaming autoregressive inference, and most of the above methods do not alleviate cache size or memory overhead.
%
% Although KV cache compression for LLMs has been widely studied, such as in StreamingLLM~\cite{xiaoefficient}, H$2$O~\cite{zhang2023h2o}, and DuoAttention~\cite{xiaoduoattention}, it remains largely unexplored in AR diffusion models.
Although KV cache compression has been widely studied in LLMs~\cite{xiaoefficient,zhang2023h2o,xiaoduoattention,zeng2026hybridkvhybridkvcache} and has been explored in autoregressive image generation~\cite{limemory,qin2025head}, it remains largely unexplored in AR video diffusion models.
To compress the KV cache of AR video diffusion models, Dummy Forcing~\cite{guo2026efficient} observes that a subset of attention heads concentrates primarily on the currently generated chunk and exploits this property for compression. 
However, it lacks a detailed characterization of the attention patterns and functional roles of individual heads, and the aggressive compression leads to discontinuities across chunks and a drop in temporal dynamics.
In contrast, we empirically identify the functional roles of different heads and perform hybrid compression based on their static and dynamic patterns, better preserving output quality.
\section{Observation}
\label{sec:observation}
In this section, we investigate the underlying principles of KV cache utilization in AR video diffusion models to motivate the compression strategy. 
We begin with intuitive observations of attention head patterns in~\Cref{sec:observation_head_pattern}, followed by empirical evidence that verifies the functional roles of different heads in~\Cref{sec:observation_head_ablation}, and finally investigate their stability and generalizability in~\Cref{sec:observation_stability}.

% To guide the design of KV compression for autoregressive video diffusion models, we conduct a series of empirical study using Self Forcing~\cite{huang2025selfforcing} and Longlive~\cite{yang2025longlive} as base models.
% Our observations are summarized as follows:

\subsection{Attention Head Pattern of Autoregressive Video Diffusion Models}
\label{sec:observation_head_pattern}

Video diffusion models typically exhibit a spatial-temporal functional specialization~\cite{xisparse,yang2025sparse}. In AR video diffusion models, the introduction of KV cache allows this property to manifest over the evolving context of autoregressive generation.. This naturally raises the following question:

\noindent \textbf{Question 1:} \textit{How do AR video diffusion models organize attention over spatiotemporal content during chunk-wise generation?}

\begin{figure}[t]
    \centering
    \includegraphics[width=\linewidth]{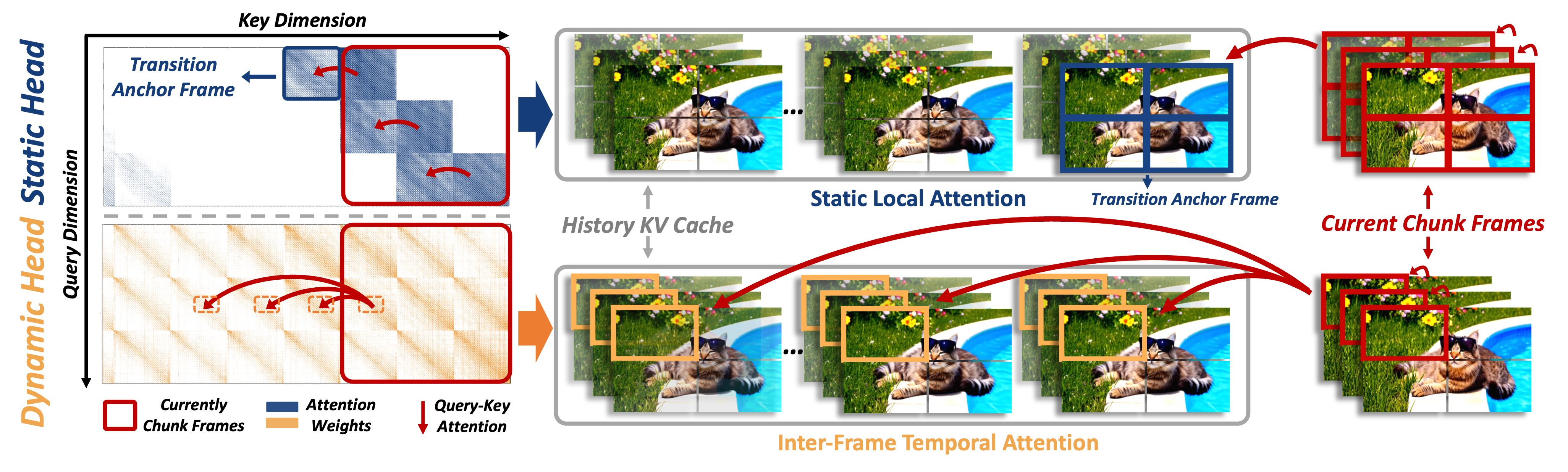}
    \caption{Attention head patterns in AR video diffusion models. Static heads focus on intra-frame dependencies and transitions across autoregressive chunks, whereas dynamic heads capture the inter-frame evolution of corresponding regions.}
    \label{fig:observation_pattern}
\end{figure}

To address this, we employ models including Wan2.1~\cite{wan2025wan}, SkyReels-V2~\cite{chen2025skyreels}, Self Forcing~\cite{huang2025selfforcing}, and Longlive~\cite{yang2025longlive} to generate videos using VBench~\cite{huang2023vbench} prompts~\footnote{We provide the detailed attention map visualization (\Cref{fig:attention_map}) in~\Cref{app:attention_pattern}.}.
Through comparison across bidirectional, autoregressive, many-step and few-step video diffusion models, we categorize the attention heads into \textbf{static head} and \textbf{dynamic head}, and summarize the patterns in~\Cref{fig:observation_pattern}:

\noindent \textbf{Observation 1 (Static and Dynamic Head Pattern):} \textit{Static heads consistently attend to the current chunk and the most recent frame, preserving intra-frame fidelity and visual continuity across local autoregressive chunks. Dynamic heads capture the inter-frame evolution of corresponding regions to exploit long-range temporal context.}

As illustrated in~\Cref{fig:observation_pattern}, the static head primarily attends to local spatial frames. Consequently, its attention map exhibits a chunk-wise pattern, with consistent attention placed on the currently generated chunk. Concurrently, static heads also place particular attention on the most recent frame in the historical cache, which we refer to as the \textit{transition anchor frame}. We regard this as a distinctive characteristic of autoregressive video diffusion models, where transitions across autoregressive chunks are primarily mediated through local, static attention to the transition anchor frame, rather than to the full set of historical frames.
Through this attention pattern, the static head provides a structural scaffold for the video. We regard it as an invariant and static behavior in autoregressive video generation, independent of the prompts and the specific generated content.

In contrast, the dynamic head exhibits a diagonal stripe pattern with a constant interval in the KV cache. This phenomenon is highly interpretable: since both the number of frames per chunk and the number of tokens per frame are fixed, the same spatial region across different frames appears with a fixed stride along the key dimension.
As a result, the dynamic head associates each generated region with information from the corresponding regions in historical frames (motion, object evolution), enabling the model to exploit long-range temporal context.
Because different spatial regions evolve dynamically over the course of the video, we refer to this head pattern as dynamic.

\subsection{Functional Properties of Static and Dynamic Heads}
\label{sec:observation_head_ablation}

Having established an intuitive interpretation of the head patterns, we proceed to further examine the functional roles of the two types of heads.

\noindent \textbf{Question 2:} \textit{What are the specific functional roles of the two types of heads, and what context in the KV cache is essential for them?}

We investigate this by conducting separate ablation studies that progressively mask the context accessible to each head until all historical frames are removed. Videos are generated using LongLive~\cite{yang2025longlive}, a high-performing model for long-horizon video generation.
For evaluation, we adopt 128 prompts from MovieGen~\citep{polyak2024movie} and use VBench-Long~\cite{huang2025vbench++} as benchmark. 
Since existing metrics do not adequately capture flickering and broken transitions at chunk boundaries, we introduce an optical-flow-based metric, termed \textit{chunk discontinuity} to measure~\footnote{The detailed formulation and metric effectiveness are provided in~\Cref{app:chunk_disc}.}.
It measures abrupt changes through the difference in optical flow between adjacent video frames.
We summarize our empirical finding as:

\noindent \textbf{Observation 2 (Functional Properties):} \textit{Static heads are crucial for visual continuity across autoregressive chunks while being insensitive to distant context. Dynamic heads govern subject consistency and motion dynamics, drawing on global context that is informative yet partially redundant.}

\begin{figure}[t]
    \centering
    \includegraphics[width=\linewidth]{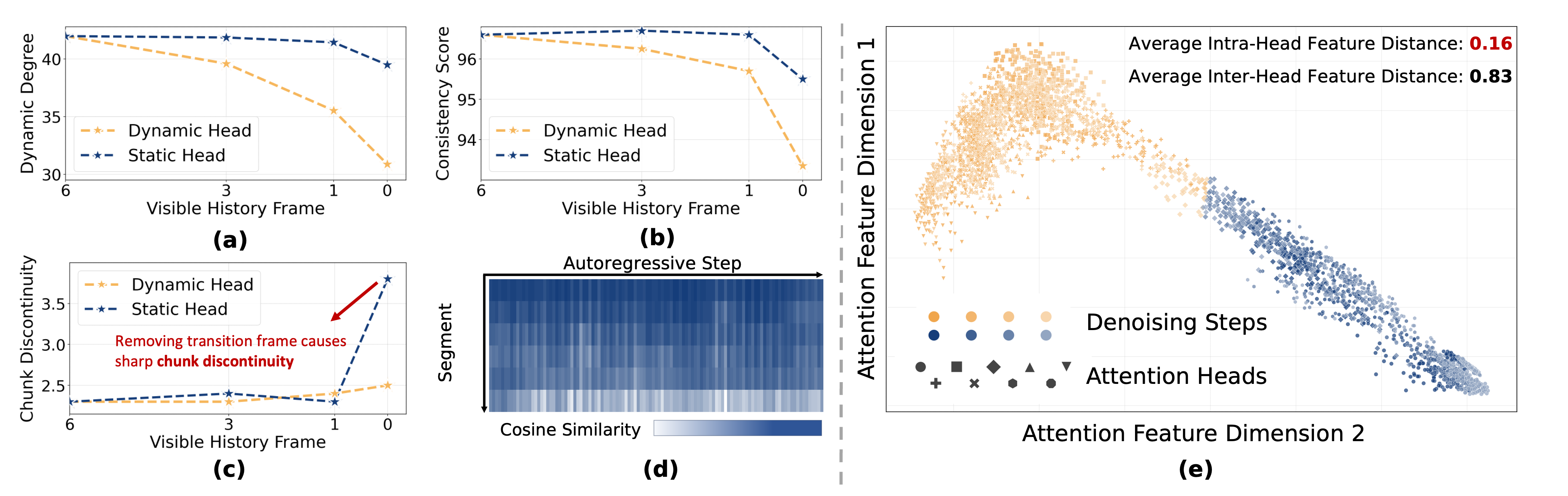}
    \caption{\textbf{Left:} (a-c) Gradually masking contextual information for dynamic heads leads to a progressive decline in dynamic degree and consistency, while masking the transition frame for static heads causes a sharp rise in chunk discontinuity, revealing different functional emphases. (d) The cosine similarity of key states of adjacent frames across different autoregressive steps and different frame segments. \textbf{Right:} (e) Principal component analysis (PCA) of attention features from a subset of attention heads, evaluated across one hundred prompt samples and four denoising steps. The observed head functioning is highly stable.}
    \label{fig:observation}
\end{figure}

As shown in \Cref{fig:observation} (a-c), as the number of visible historical frames is progressively reduced, both dynamic degree and consistency score gradually decline for dynamic heads, while remaining nearly unchanged for static heads. 
By contrast, masking the most recent frame (transition anchor frame) causes a sharp increase in chunk discontinuity for static heads, indicating significantly more abrupt transitions at chunk boundaries. We hypothesize that this effect further leads to degradation in other metrics. 
The above experiments also verify that transitions across autoregressive chunks are primarily mediated through attention to the transition anchor frame, rather than the full historical context.
% In~\Cref{sec:ablation_study}, we further examine the influence of the transition anchor frame on additional performance metrics.
%

Moreover, we observe that adjacent frames in autoregressive generation exhibit substantial regional similarity (potentially redundant), with generally high KV cache similarity that varies across different frame segments, as shown in \Cref{fig:observation} (d).
These insights provide empirical support for our hybrid compression scheme, that static heads are pruned statically while dynamic heads are pruned based on similarity, decoupling static local patterns from dynamic context utilization.

\subsection{Stability of Head Properties}
\label{sec:observation_stability}

Furthermore, we conduct a statistical analysis of the above head properties to address the question:

\noindent \textbf{Question 3:} \textit{Do the head properties remain stable, or do they exhibit substantial variation?}

To provide a comprehensive study, we experiment on LongLive with 100 standard VBench~\cite{huang2023vbench} prompts across all four denoising steps.
For a random subset of heads, we extract the key states of each latent frame in the KV cache and compute frame-wise attention features. Based on the features, we visualize the distribution using principal component analysis (PCA) as shown in~\Cref{fig:observation} (e).

\noindent \textbf{Observation 3 (Stability of Head Properties):} \textit{Head functional specialization remains stable across samples and denoising steps in its attention patterns.}

As shown in \Cref{fig:observation}(e), the features of each head form tightly clustered distributions across different samples and denoising steps, with average intra-head divergence (0.16) substantially smaller than average inter-head divergence(0.83). This provides a basis for effective head classification.

\noindent \textbf{Discussion (Autoregressive Distinctiveness):} Prior studies on bidirectional models also identify spatial-temporal head patterns~\cite{xisparse}. Our observations differ in three important aspects. First, we uncover a unique dependency on transition anchor frames that is specific to autoregressive generation. Second, our observation is grounded in the KV cache, characterizing how the query chunk attends to previously generated chunks rather than fully bidirectional attention.
Third, our compression scheme is based on temporal similarity in the KV cache rather than relying on a sparse attention pattern.

%
% This consistency suggests that attention heads in autoregressive video diffusion models have developed stable functional specialization, and can therefore be reliably classified.

% \noindent \textbf{Discussion:} \textit{How head roperties emerge in autoregressive video diffusion models?}

% To examine the generality of the above observation, we conduct experiments on Wan2.1~\cite{wan2025wan}, SkyReels-V2~\cite{chen2025skyreels}, Longlive~\cite{yang2025longlive}, and Self Forcing~\cite{huang2025selfforcing}. The detailed attention maps are provided in the Appendix. We find that distinct attention patterns over inter-frame and intra-frame regions are a \textbf{shared feature} across bidirectional, autoregressive, many-step, and few-step video diffusion models, which aligns with previous investigations ~\cite{xisparse}. We regard it as a plausible phenomenon, given that mainstream autoregressive video models inherit certain properties from their bidirectional teacher models.
% %
% However, we note that autoregressive video models evolve its \textbf{distinctive features}. First, they learn smooth transitions between the preceding frame and the current frame. 
% Second, at each generation step, autoregressive video models condition on previously generated frames, which may contain substantial redundancy. Unlike bidirectional models, autoregressive models do not require equally long query and key states, making KV cache compression over these redundant frames a feasible design choice.

\begin{figure}[t]
    \centering
    \includegraphics[width=0.85\linewidth]{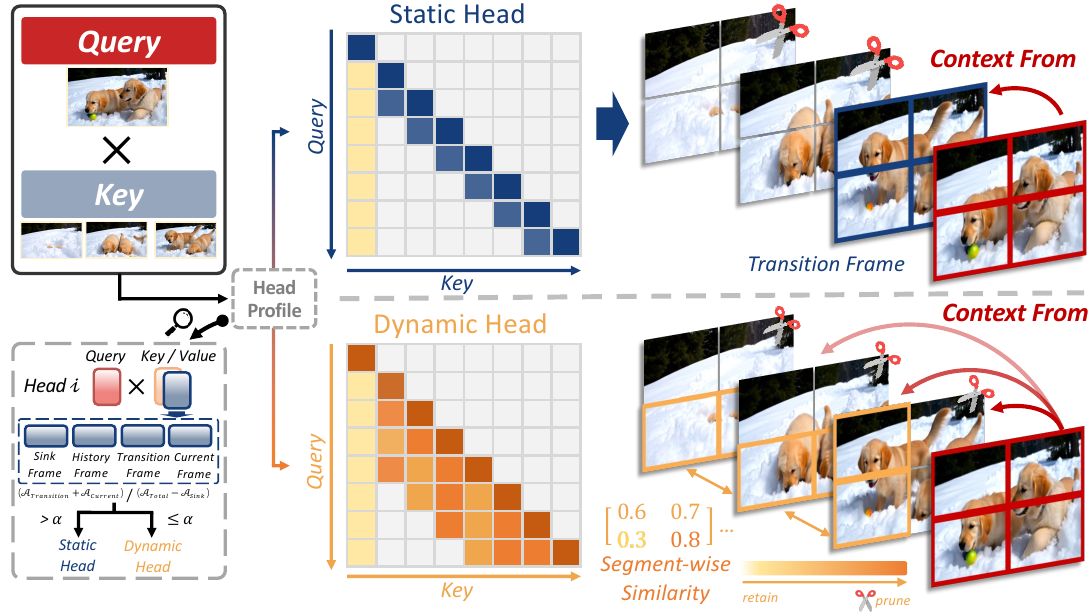}
    \caption{Overview of \method. We perform offline head profiling to classify attention heads into \textcolor{MethodBlue}{Static} and \textcolor{MethodOrange}{Dynamic}. During inference, static heads are pruned leveraging the structural pattern, while dynamic heads are pruned adaptively based on segment-wise similarity of adjacent frames. For simplicity, we use one frame per chunk as an example.}
    \label{fig:method}
\end{figure}

\section{\method}
\label{sec:method}
Motivated by the observations, we propose \method, a hybrid compression scheme for autoregressive diffusion models, as depicted in~\Cref{fig:method}.
We conduct a model-level \textbf{offline head profiling} in~\Cref{sec:method_head_profile} that identifies static and dynamic heads.
We then apply \textbf{static structural pruning} for static heads in~\Cref{sec:method_static_compression} and \textbf{dynamic similarity pruning} for dynamic heads in~\Cref{sec:method_dynamic_compression}.

\subsection{Offline Head Profiling}
\label{sec:method_head_profile}
Given the consistent functional behaviors of static and dynamic heads in \textit{Observation 3}, we propose an offline head profiling strategy to categorize them before actual inference.
According to the head pattern, the attention mass of static heads along the key dimension is concentrated on the currently generated chunk and the transition anchor frame, whereas the attention mass of dynamic heads is distributed more evenly across the entire attention window.
This provides an intuitive criterion for head classification: utilizing the proportion of total attention mass assigned to the local static frames.
Since some models apply special treatment to sink frames in their training recipes~\cite{yang2025longlive,lu2025reward,yesiltepe2025infinity}, we exclude the sink frames from this computation.
Finally, given the per-head attention mass assigned to the entire attention window $\mathcal{A}_{Total}$, the generated chunk $\mathcal{A}_{Generate}$, the transition frame $\mathcal{A}_{Transition}$, and the sink frame $\mathcal{A}_{Sink}$, the head profiling metric is defined as:
\begin{equation}
\label{eq:head_profile}
\mathrm{HeadType}=
\begin{cases}
\mathrm{\textcolor{MethodBlue}{Static}}, & \text{if } \displaystyle \frac{\mathcal{A}_{\mathrm{Generate}} + \mathcal{A}_{\mathrm{Transition}}}{\mathcal{A}_{\mathrm{Total}} - \mathcal{A}_{\mathrm{Sink}}} > \alpha,\\[6pt]
\mathrm{\textcolor{MethodOrange}{Dynamic}}, & \text{otherwise}.
\end{cases}
\end{equation}
Here, $\alpha$ is a model-specific hyperparameter, and the classification can be completed within a single prompt.
Notably, the metric aligns naturally with our subsequent compression strategy, where frames with lower accumulated attention mass are better eviction candidates, consistent with KV eviction methods such as H$_2$O~\cite{zhang2023h2o}.
In~\Cref{sec:ablation_study}, we show that this simple criterion is sufficient to distinguish the majority of heads and is not sensitive to $\alpha$, which promotes scalability.

\subsection{Static Structural Pruning for Static Heads}
\label{sec:method_static_compression}
% \label{sec:method_compression}
% In \textit{Observation 1} and \textit{Observation 2}, we show that static heads attend to fixed regions of the KV cache, whereas the importance of the cache for dynamic heads varies as generation progresses. This naturally motivates a hybrid design that compresses the two types of heads separately, decoupling static local patterns from dynamic context utilization.

In \textit{Observation 1}, we show that static heads are highly sensitive to the transition anchor frame while underutilizing distant context. 
% As shown in~\Cref{fig:observation} (c), masking distant frames in the KV cache causes almost no noticeable degradation in dynamic degree or overall consistency. By contrast, masking the most recent frame leads to pronounced discontinuities across chunks, which further results in performance drops on other metrics.
%
Therefore, we adopt a structured compression strategy for static heads by retaining the key and value states of the transition anchor frame and the current chunk to preserve intra-frame spatial structure and local chunk transitions. 
Given that each chunk contains $C$ frames and each frame consists of $F$ tokens, for the \textit{i}-th AR step, the self-attention is formulated as:
\begin{equation}
\label{eq:static_compression}
\begin{aligned}
\mathbf{O}^{\mathrm{static}}_i
& = {}  \mathrm{Attention}\Big(
Q_{iCF:(i+1)CF}, \\
& \left[ K_{sink},\; K_{(iC-1)F:iCF},\; K_{iCF:(i+1)CF} \right], 
\left[ V_{sink},\;V_{(iC-1)F:iCF},\; V_{iCF:(i+1)CF} \right]
\Big),
\end{aligned}
\end{equation}
where $Q$, $K$, and $V$ denote the query, key, and value states, and $K_{sink}$ denotes the key states of the sink frames.
This formulation statically preserves the sink frames and the transition anchor frame for autoregressive chunks, and can be readily extended to frame-wise generation models as well.

\subsection{Dynamic Similarity Pruning for Dynamic Heads}
\label{sec:method_dynamic_compression}

Dynamic heads assign high attention mass to regions separated by fixed intervals, corresponding to the same spatial locations across different frames. 
However, these segments differ substantially in their temporal evolution as shown in~\Cref{fig:observation} (d): some remain highly similar across frames with only limited variation (potentially background regions or static objects), whereas others undergo continuous changes due to motion, actions, or object evolution.
Accordingly, we assess the redundancy of different segments during generation for dynamic compression. 

Specifically, we first partition each latent frame into $n$ segments and compute the segment-wise cosine similarity between corresponding segments in each frame and its next adjacent frame. 
Given a compression ratio $r$, we evict the top-$k$ segments in each frame with the highest similarity values, preserving $(1-r)\%$ of all frame segments.
Similar to~\Cref{eq:static_compression}, let $K_{t,j}$ and $V_{t,j}$ denote the key and value state of the $j$-th segment in the $t$-th history frame, and we formulate the compression:
\begin{equation}
\label{eq:dynamic_compression1}
s_t^{(j)}
=
\operatorname{Cosine \ }\!\bigl(
K_{t,j},\,
K_{t+1,j}
\bigr),
\qquad
\mathcal{I}_t^{\mathrm{keep}}
=
\operatorname{BottomK}\!\left(
\{s_t^{(j)}\}_{j=1}^{n},\,
\lfloor rn \rfloor
\right).
\end{equation}
Here, $s_t^{(j)}$ are cosine similarity values, and $\mathcal{I}_t^{\mathrm{keep}}$ are the indices of selected low-similarity segment to keep. Notably, for segment-wise similarity computation, we use only the key states over attention heads in the first block of the diffusion transformer~\cite{Peebles2022DiT} as a proxy, thereby avoiding the substantial computation of all blocks.
We denote the compressed key and value states as $\widetilde{K}_{\mathcal H}
=
\bigcup_{t\in\mathcal H}
\{K_{t,j}\mid j\in\mathcal I_t^{\mathrm{keep}}\}$ and $\widetilde{V}_{\mathcal H}
=
\bigcup_{t\in\mathcal H}
\{V_{t,j}\mid j\in\mathcal I_t^{\mathrm{keep}}\}$, and the self-attention for dynamic head is denoted:
\begin{equation}
\label{eq:dynamic_compression3}
\begin{aligned}
\mathbf{O}^{\mathrm{dynamic}}_i
= {} & \mathrm{Attention}\Bigl(
Q_{iCF:(i+1)CF}, \\
& \left[K_{\mathrm{sink}},\, \widetilde{K}_{\mathcal H},\, K_{iCF:(i+1)CF}\right], 
\left[V_{\mathrm{sink}},\, \widetilde{V}_{\mathcal H},\, V_{iCF:(i+1)CF}\right]
\Bigr).
\end{aligned}
\end{equation}
This design is motivated by the observation that adjacent frames in a video are often similar and therefore contain redundancy within a short temporal window~\cite{gao2025ca2}. As a result, removing highly similar segments introduces only minimal information loss. In~\Cref{sec:ablation_study}, we empirically show that such a scheme offers an advantage in temporal dynamics over random and uniform token reduction.

% % \subsection{Other Optimization}
% \subsection{Seamless Integration of Quantization}
% \label{sec:method_optimization}

% We further incorporate FP8 quantization~\cite{zhang2024sageattention2}
% into our KV cache compression method. This optimization reduces the computational cost of attention module by leveraging FP8 attention kernels tailored to the NVIDIA Hopper architecture, which further boosts throughput with minimal performance drop, as shown in~\Cref{tab:main_result_long}. This shows that our method is compatible with other acceleration techniques.

% \paragraph{Rolling Cache.}
% Previous method~\cite{guo2026efficient} adopt a naive rolling cache that updates the context by explicitly shifting historical key-value tokens and appending new ones, which introduces repeated memory copies at each step. 
% %
% We replace this design with a fixed-capacity ring buffer that maintains a write pointer to reduce memory traffic.

\section{Experiments}
\label{sec:experiment}

% \subsection{Setup}
\paragraph{Models and Baselines.}
We conduct the experiments using mainstream AR video generation models including Self Forcing~\citep{huang2025selfforcing} and  Longlive~\citep{yang2025longlive}~\footnote{Results on Krea-Realtime-14B~\cite{erwann2025krea} and interactive video generation on Longlive are in~\Cref{app:krea_14B,app:interactive_result}.}.
% For Longlive, both single and interactive prompt are evaluated.
We compare our method against both the full KV cache setting and representative KV cache compression baselines. StreamingLLM~\cite{xiaoefficient} serves as a naive baseline that uniformly retains sink and recent frames for all heads. Dummy Forcing~\citep{guo2026efficient} employs an aggressive local pruning strategy.
We provide method implementation
details in~\Cref{app:implementation_detail}.

\paragraph{Benchmarks and Evaluation Metrics.}
We evaluate both short and long video generation on VBench~\citep{huang2023vbench} and VBenchLong~\citep{huang2025vbench++}, and conduct a user study~\footnote{Setup, protocol, and screenshots of the user study are in~\Cref{app:user_study}.}. Specifically, we generate 5-second videos using 946 official VBench prompts and evaluate all 16 dimensions. For 30-second and 60-second videos, we adopt 128 prompts from MovieGen~\citep{polyak2024movie}, weighting the total score using the standard VBench coefficients, consistent with previous work~\citep{yin2025slow,lu2025reward}. All prompts are sampled with 5 different seeds.
%
% Similar to~\citep{lu2025reward,yuan2026helios}, we compute the dynamic degree of the last 5-second clips to measure drift in long videos, which we denote as \textit{tail dynamic}.
To quantify the continuity of chunk transition, we use \textit{chunk discontinuity} metric as defined in~\Cref{app:chunk_disc}.
For the efficiency metric, we calculate the frames generated per second (FPS) within the diffusion transformer (DiT) and the corresponding speedups on a single NVIDIA H200 GPU, together with the GPU memory usage of KV cache. 
% We exclude the VAE overhead since it is not the main bottleneck, and can be properly hidden through pipelining~\cite{low2025talkingmachines,huang2025live}.

\begin{table*}[t]
\centering
\renewcommand{\arraystretch}{0.95}
\setlength{\tabcolsep}{5pt}

{\tiny
$\uparrow$ indicates higher is better, $\downarrow$ indicates lower is better.
\textcolor{gainblue}{\ding{115}} highlights improved performance over Full KV.
}

\resizebox{\textwidth}{!}{%
\begin{tabular}{c|lcc|cc|cccccc}
\toprule
\multicolumn{1}{c|}{\multirow{2}{*}[-0.8em]{\makecell[c]{\textbf{Setting}}}}
& \multicolumn{1}{c}{\multirow{2}{*}[-0.8em]{\makecell[c]{\textbf{Method}}}}
& \multicolumn{2}{c|}{\textbf{Efficiency Metrics}}
& \multicolumn{2}{c|}{\textbf{Specific Metrics}}
& \multicolumn{6}{c}{\textbf{General Metrics}} \\
\cmidrule(lr){3-4}
\cmidrule(lr){5-6}
\cmidrule(lr){7-12}
&
& FPS$\uparrow$
& Speedup$\uparrow$
& \makecell[c]{Chunk\\Disc.$\downarrow$}
& \makecell[c]{Dynamic\\Degree$\uparrow$}
& \makecell[c]{\textbf{Total}\\ \textbf{Score}$\uparrow$}
& \makecell[c]{Imaging\\Quality$\uparrow$}
& \makecell[c]{Subject\\Cons.$\uparrow$}
& \makecell[c]{Background\\Cons.$\uparrow$}
& \makecell[c]{Motion\\Smooth.$\uparrow$}
& \makecell[c]{Aesthetic\\Quality$\uparrow$} \\
\midrule

\multirow{5}{*}{\rotatebox{90}{\makecell[c]{LongLive\\60-second}}}
& \gray{Full KV}
& \gray{20.48}
& \gray{1.00$\times$}
& \gray{2.6}
& \gray{42.40}
& \gray{80.23}
& \gray{68.84}
& \gray{97.82}
& \gray{96.82}
& \gray{98.77}
& \gray{61.47} \\

& StreamingLLM
& 22.50 & 1.10$\times$ & 2.5 & 40.89
& 80.28\gainnote & 69.40 & 97.91 & 96.88 & 98.75 & 61.78 \\

& Dummy Forcing ($L=1$)
& 28.06 & 1.37$\times$ & 3.6 & 26.02
& 79.35 & 71.25 & 97.95 & 96.94 & 98.79 & 62.01 \\

& Dummy Forcing ($L=2$)
& 26.36 & 1.29$\times$ & 2.9 & 34.10
& 79.44 & 70.71 & 97.38 & 96.52 & 98.45 & 61.40 \\

& \orangecell{\textbf{\method\ (Ours)}}
& \orangecell{26.71}
& \orangecell{1.30$\times$}
& \orangecell{2.5\gainnote}
& \orangecell{43.56\gainnote}
& \orangecell{\textbf{80.43}\gainnote}
& \orangecell{70.24}
& \orangecell{97.79}
& \orangecell{96.66}
& \orangecell{98.57}
& \orangecell{61.50} \\
\midrule

\multirow{5}{*}{\rotatebox{90}{\makecell[c]{LongLive\\30-second}}}
& \gray{Full KV}
& \gray{21.10}
& \gray{1.00$\times$}
& \gray{2.4}
& \gray{42.54}
& \gray{80.38}
& \gray{68.91}
& \gray{97.99}
& \gray{96.93}
& \gray{98.80}
& \gray{61.67} \\

& StreamingLLM
& 22.34 & 1.06$\times$ & 2.4 & 40.93
& 80.38 & 69.35 & 96.06 & 96.98 & 98.78 & 61.92 \\

& Dummy Forcing ($L=1$)
& 27.45 & 1.30$\times$ & 3.0 & 26.56
& 79.37 & 70.73 & 98.07 & 96.99 & 98.82 & 62.06 \\

& Dummy Forcing ($L=2$)
& 26.21 & 1.24$\times$ & 2.3 & 33.90
& 79.61 & 70.65 & 97.64 & 96.64 & 98.54 & 61.73 \\

& \orangecell{\textbf{\method\ (Ours)}}
& \orangecell{26.77}
& \orangecell{1.27$\times$}
& \orangecell{2.4}
& \orangecell{43.65\gainnote}
& \orangecell{\textbf{80.65}\gainnote}
& \orangecell{70.29}
& \orangecell{98.00}
& \orangecell{96.81}
& \orangecell{98.66}
& \orangecell{61.98} \\
\midrule

\multirow{5}{*}{\rotatebox{90}{\makecell[c]{Self Forcing\\30-second}}}
& \gray{Full KV}
& \gray{17.76}
& \gray{1.00$\times$}
& \gray{3.4}
& \gray{46.86}
& \gray{79.72}
& \gray{67.63}
& \gray{97.20}
& \gray{96.38}
& \gray{98.14}
& \gray{61.14} \\

& StreamingLLM
& 21.86 & 1.23$\times$ & 2.8\gainnote & 54.50\gainnote
& 80.06\gainnote & 67.90 & 96.93 & 96.16 & 98.10 & 59.67 \\

& Dummy Forcing ($L=1$)
& 27.75 & 1.56$\times$ & 3.5 & 46.55
& 79.95\gainnote & 69.05 & 97.20 & 96.38 & 98.14 & 61.13 \\

& Dummy Forcing ($L=6$)
& 22.11 & 1.24$\times$ & 3.4 & 50.47\gainnote
& 79.78\gainnote & 68.83 & 96.62 & 96.03 & 98.03 & 59.97 \\

& \orangecell{\textbf{\method\ (Ours)}}
& \orangecell{26.65}
& \orangecell{1.50$\times$}
& \orangecell{2.7\gainnote}
& \orangecell{52.23\gainnote}
& \orangecell{\textbf{80.07\gainnote}}
& \orangecell{68.67}
& \orangecell{97.00}
& \orangecell{96.08}
& \orangecell{98.08}
& \orangecell{60.16} \\

\bottomrule
\end{tabular}%
}

\caption{Quantitative results on efficiency and quality on \textbf{long video generation} with VBenchLong.}
\label{tab:main_result_long}
\end{table*}

\begin{figure*}[t]
\captionsetup{justification=raggedright,singlelinecheck=false}
\centering

\begin{minipage}[t]{0.71\textwidth}
\vspace{0pt}
\captionsetup{justification=raggedright,singlelinecheck=false}
\renewcommand{\arraystretch}{0.94}
\setlength{\tabcolsep}{4.5pt}
\centering

\resizebox{\linewidth}{!}{%
\begin{tabular}{c|lcc|cc|ccc}
\toprule
\multicolumn{1}{c|}{\multirow{2}{*}[-0.7em]{\makecell[c]{\textbf{Setting}}}}
& \multicolumn{1}{c}{\multirow{2}{*}[-0.7em]{\makecell[c]{\textbf{Method}}}}
& \multicolumn{2}{c|}{\textbf{Efficiency Metrics}}
& \multicolumn{2}{c|}{\textbf{Specific Metrics}}
& \multicolumn{3}{c}{\textbf{General Metrics}} \\
\cmidrule(lr){3-4}
\cmidrule(lr){5-6}
\cmidrule(lr){7-9}
&
& FPS$\uparrow$
& Speedup$\uparrow$
& \makecell[c]{Chunk\\Disc.$\downarrow$}
& \makecell[c]{Dynamic\\Degree$\uparrow$}
& \makecell[c]{\textbf{Total}\\\textbf{Score}$\uparrow$}
& \makecell[c]{Quality\\Score$\uparrow$}
& \makecell[c]{Semantic\\Score$\uparrow$} \\
\midrule

\multirow{5}{*}{\rotatebox{90}{\makecell[c]{LongLive\\5-second}}}
& \gray{Full KV}
& \gray{21.85}
& \gray{1.00$\times$}
& \gray{2.1}
& \gray{40.28}
& \gray{83.19}
& \gray{83.63}
& \gray{81.41} \\

& StreamingLLM
& 24.56 & 1.12$\times$ & 2.1 & 43.33\gainnote & 82.85 & 83.29 & 81.09 \\

& Dummy Forcing ($L=1$)
& 29.69 & 1.36$\times$ & 2.4 & 37.22 & 82.83 & 83.33 & 80.86 \\

& Dummy Forcing ($L=2$)
& 28.36 & 1.30$\times$ & 2.1 & 36.67 & 83.15 & 83.73 & 80.85 \\

& \orangecell{\textbf{\method\ (Ours)}}
& \orangecell{29.58}
& \orangecell{1.35$\times$}
& \orangecell{2.1}
& \orangecell{45.56\gainnote}
& \orangecell{\textbf{83.23}\gainnote}
& \orangecell{83.84}
& \orangecell{80.80} \\
\midrule

\multirow{5}{*}{\rotatebox{90}{\makecell[c]{Self Forcing\\5-second}}}
& \gray{Full KV}
& \gray{19.56}
& \gray{1.00$\times$}
& \gray{2.1}
& \gray{66.39}
& \gray{83.91}
& \gray{84.71}
& \gray{80.70} \\

& StreamingLLM
& 23.36 & 1.19$\times$ & 2.4 & 65.00 & 83.80 & 84.52 & 80.89 \\

& Dummy Forcing ($L=1$)
& 28.31 & 1.45$\times$ & 2.6 & 64.44 & 83.87 & 84.64 & 80.81 \\

& Dummy Forcing ($L=6$)
& 25.41 & 1.30$\times$ & 2.4 & 63.33 & 83.79 & 84.50 & 80.96 \\

& \orangecell{\textbf{\method\ (Ours)}}
& \orangecell{28.18}
& \orangecell{1.44$\times$}
& \orangecell{2.1}
& \orangecell{69.17\gainnote}
& \orangecell{\textbf{83.98}\gainnote}
& \orangecell{84.82}
& \orangecell{80.61} \\
\bottomrule
\end{tabular}%
}
\captionof{table}{Quantitative results on \textbf{short video generation} with VBench.}
\label{tab:main_result_short}
\end{minipage}
\hfill
\begin{minipage}[t]{0.28\textwidth}
\vspace{0pt}
\centering
\captionsetup{justification=centering}
\includegraphics[width=0.97\linewidth]{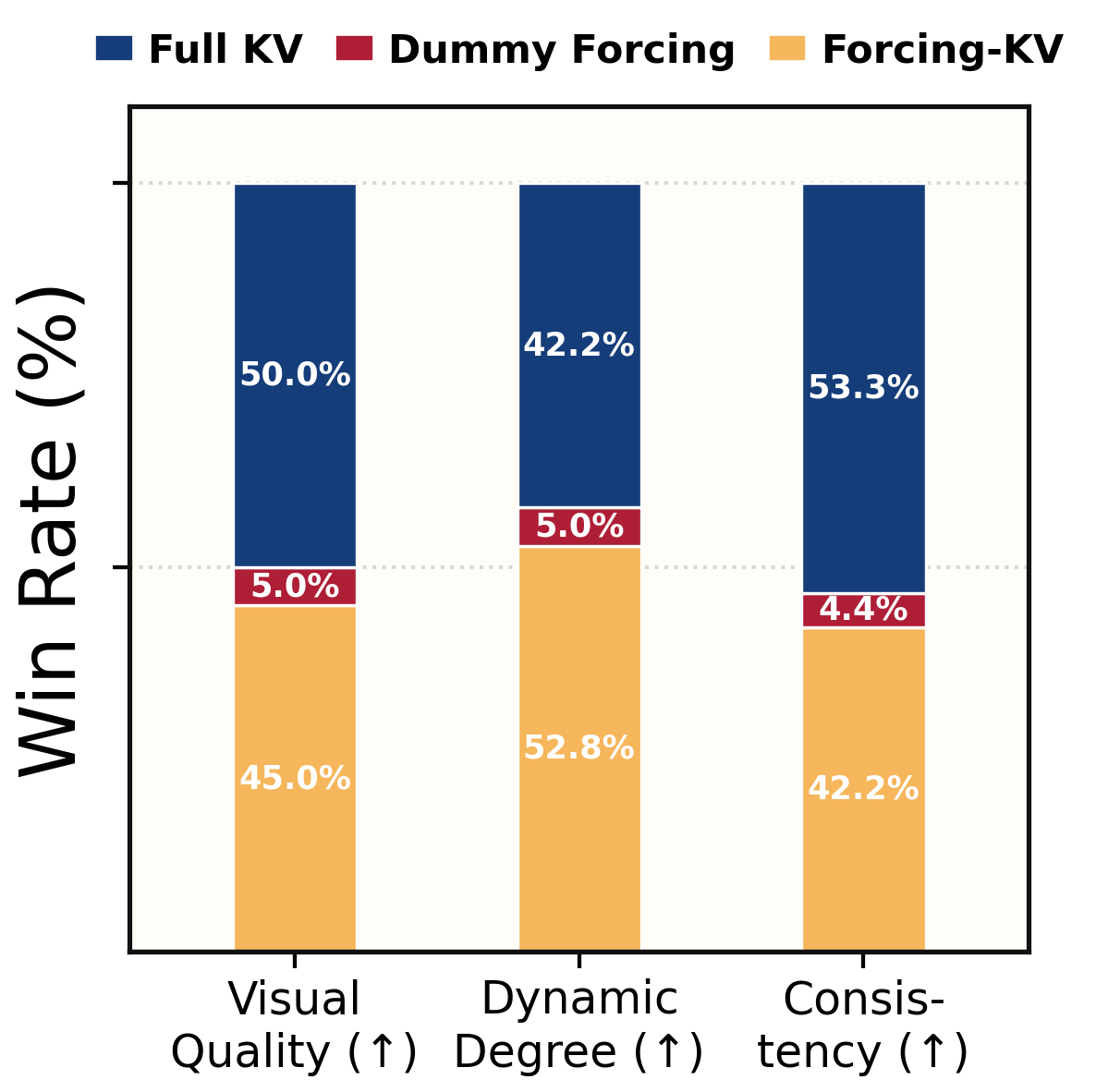}
\vspace{-5pt}
\captionof{figure}{User study.}
\label{fig:win_rate}
\end{minipage}

\end{figure*}

% \ding{172}
\subsection{Main Results}
\paragraph{Comparison with Full KV Cache.} 
As shown in~\Cref{tab:main_result_long,tab:main_result_short} and~\Cref{fig:win_rate}, we evaluate our method on both LongLive and Self Forcing across 5-second, 30-second and 60-second videos.
Our method achieves 1.30$\times$ and 1.50$\times$ inference speedups on LongLive and Self Forcing for long video generation, and 1.35$\times$ and 1.44$\times$ speedups for short video generation. 
% Short generation is generally faster because the attention window is smaller during early steps.
Through our head-wise hybrid cache compression, only $\sim$27\% and $\sim$46\% of the KV cache participate in self-attention computation for Self Forcing and Longlive.
At the same time, \method maintains comparable or slightly improved performance on VBench (80.43 vs. 80.23). In the user study, \method also achieves comparable visual quality (45.0\% vs. 50.0\%) and stronger temporal dynamics (52.8\% vs. 42.2\%). We assume that most existing base models treat all heads uniformly, forcing certain heads (i.e., static heads) to attend to distant context, but such capability may not be sufficiently learned. As a result, compressing the context for these heads can even lead to a positive effect. This also explains why StreamingLLM can maintain the performance even after evicting partial distant tokens. The results show the presence of redundancy in KV cache utilization, lending support to our observations.

\paragraph{Comparison with Other Compression Strategies.} 
% Our method demonstrates the best efficiency-performance trade-off. 
\ding{172} Vs. StreamingLLM: 
With its sliding window design closely aligned with the training regime of the base model, StreamingLLM serves as a competitive baseline. However, \method achieves head-level decomposition, which leads to substantially higher compression ratios and resulting speedups (1.50$\times$ vs. 1.23$\times$).
\ding{173} Vs. Dummy Forcing: 
While achieving comparable speedups, \method consistently achieves substantially much higher quality metrics. 
Specifically, our compression is grounded in the observations from~\Cref{sec:observation} and preserves the transition anchor frame, whereas Dummy Forcing does not. This results in markedly lower chunk discontinuity (2.5 vs. 3.6 and 2.7 vs. 3.5). Moreover, the aggressive compression of Dummy Forcing leads to a pronounced degradation in dynamic degree (26.02 vs. 43.56), even under conservative settings (34.10) when $L=6$. 
Though it achieves higher image quality in some cases, this may reflect the benchmark's preference for static content.
Notably, chunk continuity and temporal dynamics have a substantial impact on perceptual quality, which also explains why our method shows a clear advantage over Dummy Forcing in the user study (45.0\% vs. 5.0\%)~\footnote{Qualitative examples are provided in~\Cref{app:quality example}.}.

% Notably, although its drop in dynamic degree on Self Forcing is not obvious, this is likely because the severe chunk discontinuity introduces cyclic changes across chunks, thereby masking the effects of excessive context compression. This phenomenon can also be observed in the quality examples provided in the \textit{Appendix}.

% \paragraph{Short Video Generation.}
% On short videos, the same trend still holds: \method continues to prevail in terms of chunk continuity, temporal dynamics, and overall score. Unlike long video generation, short video generation (5s) terminates just as the sliding window reaches its maximum size, which prevents the long-term effects of acceleration methods and error accumulation from becoming fully apparent. Nevertheless, \method still exhibits lower chunk discontinuity than Dummy Forcing (2.1 vs. 2.6) in the short term, demonstrating its generalizability under short window settings.

\subsection{Scaling Law for Attention Window Size and Resolution}
\begin{figure}[t]
    \captionsetup{skip=0pt}
    \centering
    \includegraphics[width=\linewidth]{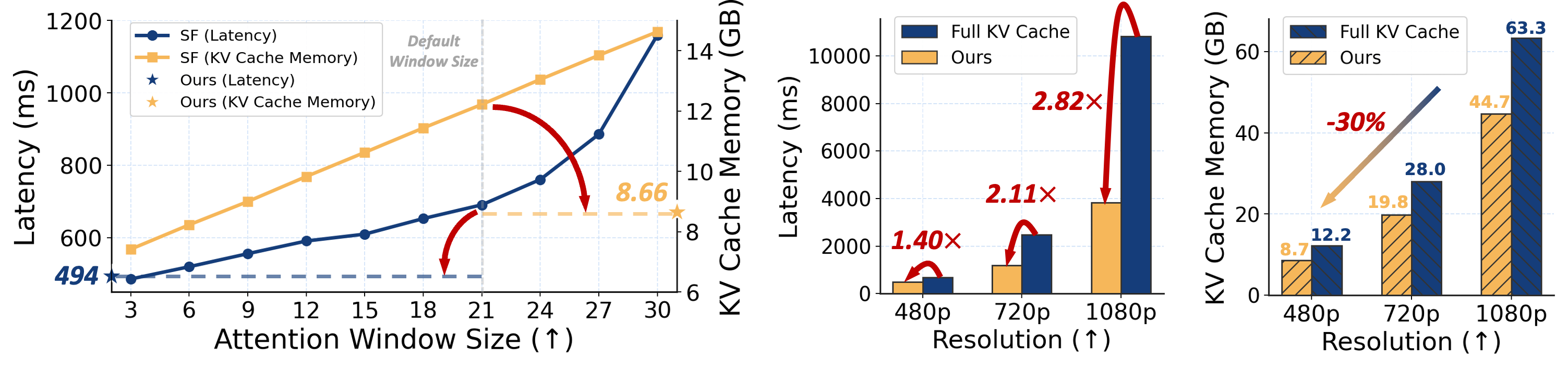}
    \caption{Scaling \method on Self Forcing with attention window size and resolution.}
    \label{fig:scaling_law}
\end{figure}

In~\Cref{tab:main_result_long,tab:main_result_short}, the speedup gains achieved by \method are bounded by the KV cache size.
However, we empirically show that the acceleration benefits of \method become increasingly pronounced as the attention window and resolution grow.
In \Cref{fig:scaling_law}, we show the latency and memory of Self Forcing.
With the attention window and video resolution increasing, the KV cache grows accordingly, causing the attention computation to scale quadratically and the memory consumption to scale linearly. 
Therefore, for the same compression ratio, the resulting speedup becomes more significant.
Under this trend, \method delivers increasing gains, rising from 1.40$\times$ to 2.82$\times$ with a memory reduction of $\sim$30\%. 
Notably, the effective KV size is even smaller since dynamic heads retain historical frames for similarity computation. A potential retrieval strategy may further reduce memory usage.
We argue that high-resolution video generation and longer video contexts are promising future directions, which further highlight the potential of our method.

\subsection{Ablation Study}
\label{sec:ablation_study}
\begin{figure}[t]
\centering

\begin{minipage}[t]{0.48\linewidth}
    \vspace{0pt}
    \captionsetup{justification=raggedright,singlelinecheck=false}
    \renewcommand{\arraystretch}{0.4}
    \setlength{\tabcolsep}{5pt}
    \raggedright
    \resizebox{\linewidth}{!}{%
    \begin{tabular}{@{}lccc@{}}
    \toprule
    Method
    & \makecell{Chunk\\Disc.$\downarrow$}
    & \makecell{Dynamic\\Degree$\uparrow$}
    & \makecell{Total\\Score$\uparrow$} \\
    \midrule

    \multicolumn{4}{c}{\textit{Head Profiling}} \\
    \midrule

    \textbf{\method ($\alpha=0.8$)}
    & 2.4 & 43.65 & 80.65 \\

    \method ($\alpha=0.5$)
    & 2.4 & 42.87 & 80.63 \\

    Random Profiling
    & 2.5 & 40.72 & 80.44 \\
    %3.7 % 79.56, 28.97,

    Human Profiling
    & 2.3 & 44.44 & 80.71 \\

    \midrule

    \multicolumn{4}{c}{\textit{KV Cache Compression}} \\
    \midrule

    \textbf{\method}
    & 2.4 & 43.65 & 80.65 \\

    w/o static-head cache
    & 4.1 & 42.58 & 79.57 \\

    w/o dynamic-head cache
    & 2.6 & 40.78 & 80.25 \\

    \bottomrule
    \end{tabular}%
    }
    \vspace{0pt}
    \captionof{table}{Ablation study of head profiling strategy and hybrid KV cache compression on LongLive (30-second video generation).}
    \label{tab:ablation_longlive_30s}
\end{minipage}
\hfill
\begin{minipage}[t]{0.49\linewidth}
    \vspace{0pt}
    \centering
    \includegraphics[width=0.95\linewidth]{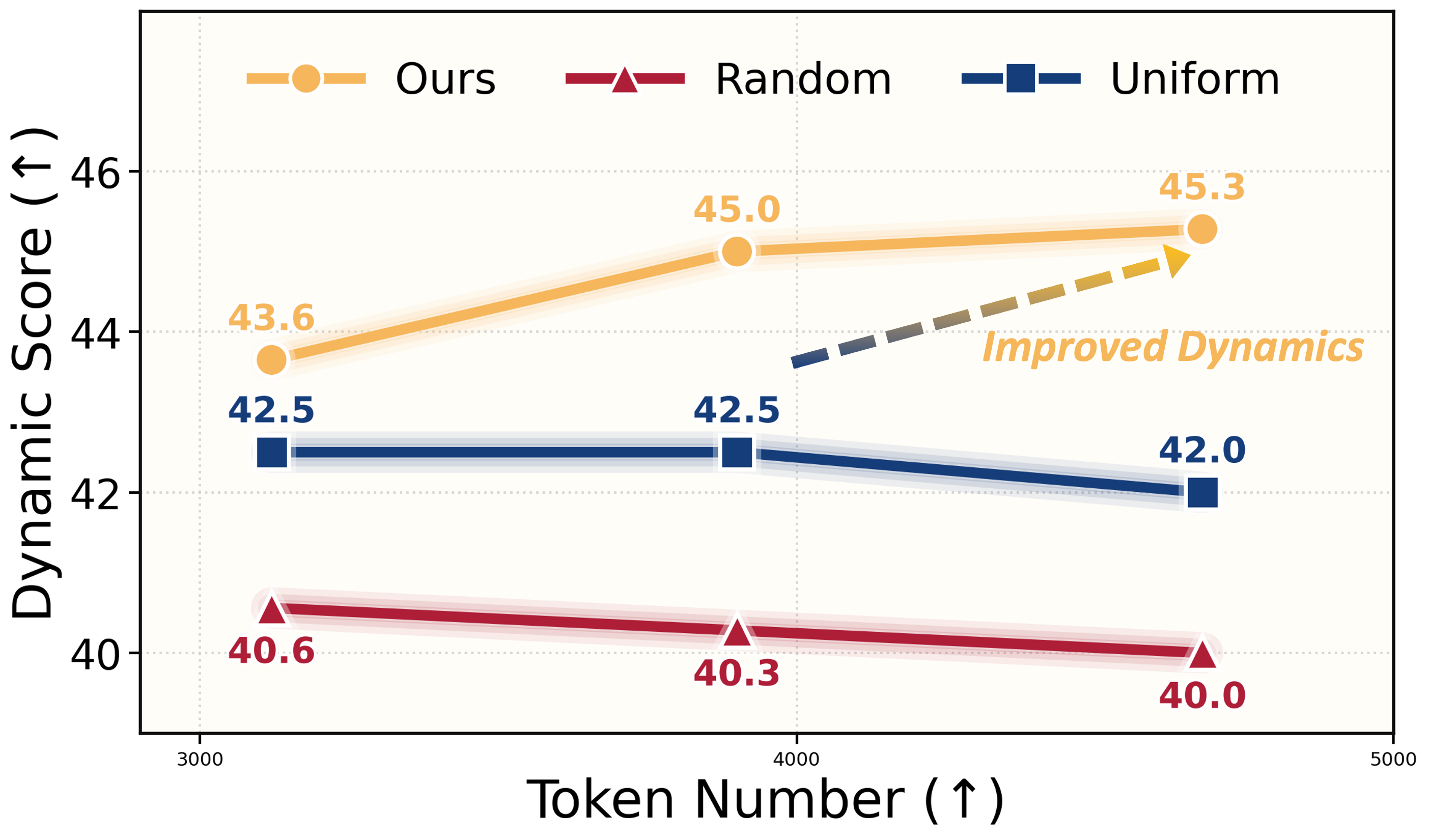}
    \caption{Dynamic score comparison of random token pruning, uniform token pruning, and our proposed similarity pruning.}
    \label{fig:pruning_criterion}
\end{minipage}

\end{figure}

In~\Cref{tab:ablation_longlive_30s}, we conduct separate ablation studies on the effectiveness of the head profiling strategy, the hybrid compression design, and the dynamic similarity pruning strategy. Unless otherwise specified, all experiments are performed on 30-second video generation with LongLive~\footnote{Additional ablation results for 5-second video generation on LongLive and Self Forcing are in~\Cref{app:more_ablation}.}.

\paragraph{Effectiveness of Head Profiling.}
Owing to the clear distinction among head types and their stability, the simple profiling strategy of \method is sufficient to identify the majority of heads and achieves performance close to that of a strong manual profiling baseline (80.65 vs. 80.71). 
We also include a random profiling baseline. 
%
% The advantage of \method on dynamic degree (43.65 vs. 40.72) over random profiling further highlights the necessity of meaningful head classification.
Because the cache retained for dynamic heads also subsumes the static portion, the degradation under random profiling mainly arises when dynamic heads are misclassified as static, causing dynamic degree to drop (40.72 vs. 43.65).
This indicates the necessity of meaningful head classification.
In addition, we find that head profiling of \method is insensitive to the hyperparameter $\alpha$. Decreasing $\alpha$ from 0.8 to 0.5 classifies more heads as static heads, which leads to only a slight drop in dynamic degree.

\paragraph{Effectiveness of Hybrid Compression.}
\method retains the transition anchor frame for static heads and preserves the cache segments for dynamic heads.
We separately prune the corresponding cache of each head type to study the effect of hybrid head modeling, denoted as \textit{w/o static-head cache} and \textit{w/o dynamic-head cache}.
As shown in~\Cref{tab:ablation_longlive_30s}, pruning the cache of static head leads to a substantial increase in chunk discontinuity (4.1 vs. 2.4), which in turn degrades the overall score. In contrast, pruning all cache of dynamic head mainly reduces the dynamic degree (40.78 vs. 43.65), consistent with our finding in~\Cref{sec:observation_head_ablation}. 
This indicates that the two components retained by our compression strategy are indeed effective, and that they correspond to their function roles.

\paragraph{Effectiveness of Dynamic Similarity Pruning.}
To validate the effectiveness of the dynamic similarity pruning strategy in \method, we conduct comparisons against other pruning criteria including random token pruning and uniform token pruning, as shown in~\Cref{fig:pruning_criterion}.
Our method consistently achieves higher dynamic degree, and improves temporal dynamics as the token budget increases. We attribute this advantage to two factors. First, adjacent frames in autoregressive generation often exhibit substantial similarity as visualized in~\Cref{fig:observation} (d), making them suited for cross-frame pruning. Second, our method operates in a segment-wise manner, which is better aligned with the continuity of video content than discrete token-wise strategies. This design is also consistent with our observation in~\Cref{fig:observation_pattern} that the stripe patterns of dynamic heads exhibit a certain width~\footnote{We integrate \method with quantization in~\Cref{app:quantization}.}.

\section{Conclusion}
\label{conclusion}

We presented \method, a hybrid KV cache compression framework for autoregressive video diffusion models.
We begin by uncovering a universal head specialization pattern shared across mainstream autoregressive video diffusion models, which naturally motivates our compression strategy.
While maintaining output quality, \method achieves a generation speed of over 29 FPS, delivering up to 1.35$\times$ and 1.50$\times$ speedups together with 30\% cache memory reduction, and scales effectively with attention window size and resolution, reaching up to 2.82$\times$ acceleration.
Our work reveals the underlying mechanisms of KV cache utilization in autoregressive video generation, providing new empirical insights and compression techniques into efficient video cache utilization.
%
% The primary limitation of this paper is that we do not investigate KV cache reduction during the training of autoregressive video models, which could potentially support longer context window.

% Meanwhile, some existing works instead approach long-video generation through video continuation~\cite{meituanlongcatteam2025longcatvideotechnicalreport,yuan2026helios}, which lies outside the scope of our method. 
% Our study moves toward the \textit{train-long-test-long} paradigm, enabling the model to condition on increasingly long contexts~\cite{zhang2025framepack,cai2026mode} more efficiently.
% Through investigating head-wise properties in KV cache, we explore how such properties can be leveraged for effective and efficient compression.

% In future work, we plan to extend this KV cache compression framework to the training stage of autoregressive models, further decoupling static attention patterns from dynamic compression to support longer context ranges with minimal memory and computational costs.

\section{Limitation and Future Works}
\label{app:limitation}
While our hybrid compression framework yields substantial gains in efficiency and performance for AR video diffusion models, it can be further improved in the following aspects. 
First, since mainstream open-source autoregressive video diffusion models are currently trained under the \textit{Self Forcing} paradigm, our observations are primarily based on existing model families, including Self Forcing~\cite{huang2025selfforcing}, Longlive~\cite{yang2025longlive}, SkyReels-V2~\cite{chen2025skyreels}, Krea-Realtime-14B~\cite{erwann2025krea} and the broader family of “forcing” models~\cite{lu2025reward,cui2025self,yesiltepe2025infinity,cui2026lol,liu2025rolling,xiao2025knot,zhu2026causal}. Nevertheless, we believe that these observations are closely tied to the fundamental principles of autoregressive generation, and that evaluating them on future autoregressive models would be an interesting direction for further study.
Second, as a training-free method, investigating KV cache reduction during the training stage of autoregressive video models is left for future work, which could potentially support longer context window through fine-tuning. 

\newpage
\bibliography{reference}
\bibliographystyle{IEEEtran}

% \begin{ack}
% Use unnumbered first level headings for the acknowledgments. All acknowledgments

% Do {\bf not} include this section in the anonymized submission, only in the final paper. You can use the \texttt{ack} environment provided in the style file to automatically hide this section in the anonymized submission.
% \end{ack}

\newpage
\newpage
\appendix
\crefalias{section}{appendix}
\Crefname{appendix}{Appendix}{Appendices}
\crefname{appendix}{appendix}{appendices}

\newpage
\section{Chunk Discontinuity}
\label{app:chunk_disc}

\paragraph{Definition of chunk discontinuity.}
In~\Cref{sec:observation} and~\Cref{sec:experiment}, we use \textit{chunk discontinuity} to quantify transitions across chunks. To ensure fairness and validity in evaluation, we define chunk discontinuity as an intuitive metric. 
Specifically, for a video containing $F$ frames and generated autoregressively in $K$ chunks, we first compute the optical flow difference between every pair of adjacent frames using \textit{Recurrent All-Pairs Field Transforms} (RAFT)~\cite{teed2020raft}, a popular deep network architecture for optical flow computation. Then, we divide the average of the Top-$(K-1)$ largest values by the overall mean. The metric is defined as:
\begin{equation}
\delta_t=\Delta\mathrm{RAFT}(I_t,I_{t+1}),\quad t=1,\ldots,F-1
\end{equation}
\vspace{-4pt}
\begin{equation}
\mathrm{Chunk \ Disc.}=
\frac{
\mathrm{Sum}\!\left(\mathrm{Top}_{K-1}\!\left(\{\delta_t\}_{t=1}^{F-1}\right)\right)/(K-1)
}{
\mathrm{Sum}\!\left(\{\delta_t\}_{t=1}^{F-1}\right)/(F-1)
}
\end{equation}

\begin{figure}[H]
    \centering
    \includegraphics[width=\linewidth]{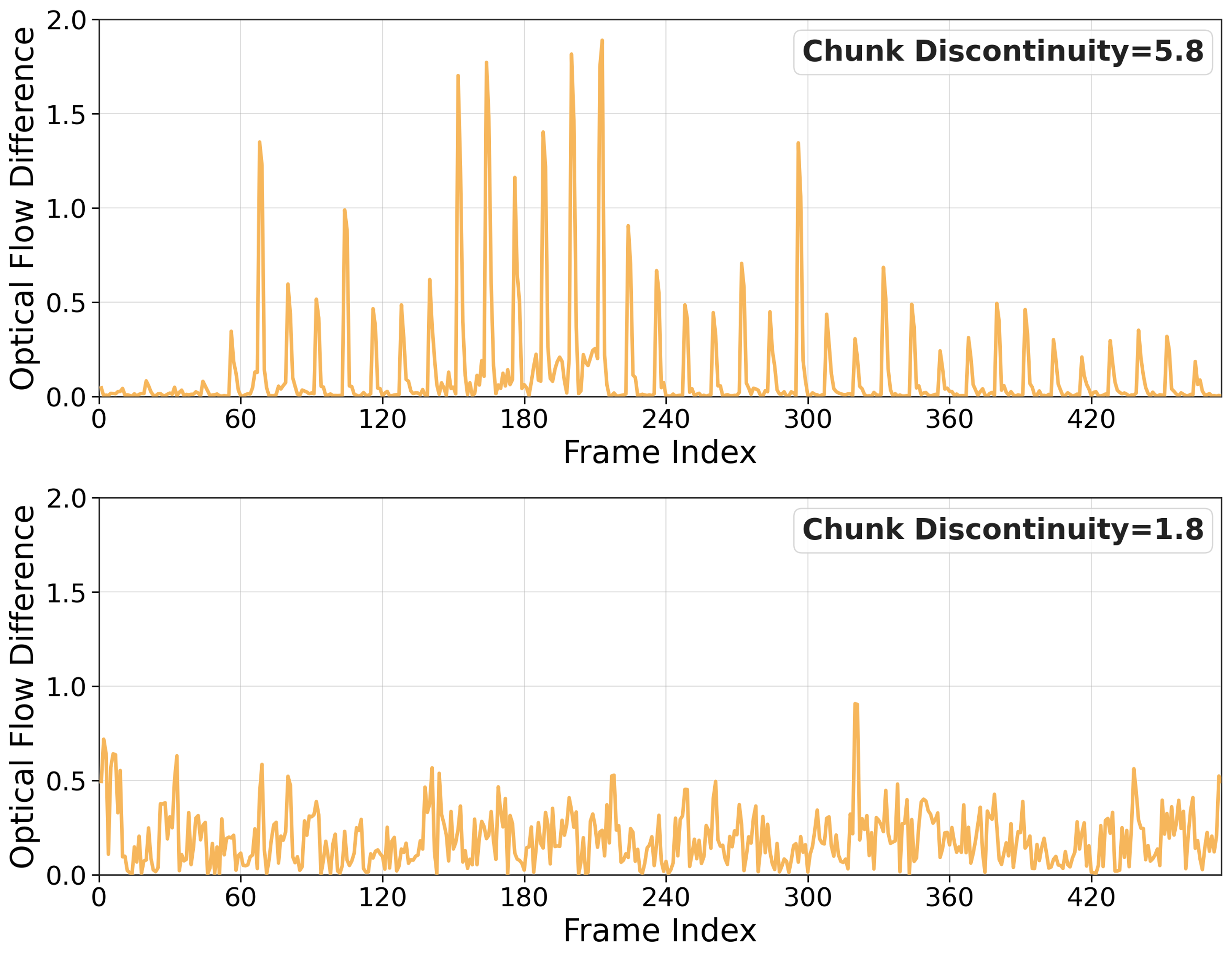}
    \caption{Case study of optical flow difference variations for a 30-second (\textasciitilde480 frames) video with high or low metric value.}
    \label{fig:optical_value_plot}
\end{figure}

\paragraph{Effectiveness of chunk discontinuity.} 
Our metric is applicable to existing autoregressive diffusion models, which generate continuous videos without scene cut. 
Under this setting, for a video that is temporally smooth and continuous, the optical flow difference between adjacent frames should vary relatively uniformly, resulting in a uniformly lower metric value. Conversely, a low metric value also indicates a lower average peak difference, suggesting milder temporal variation in the video.
In~\Cref{fig:optical_value_plot}, we provide an example of a video with a high metric value and poor chunk continuity. The local maxima can be seen to appear regularly at fixed-interval chunk boundaries, demonstrating that our metric effectively captures discontinuities across chunks.

\newpage
\section{Attention Patterns of Various Diffusion Models}
\label{app:attention_pattern}

\begin{figure}[H]
    \centering
    \includegraphics[width=\linewidth]{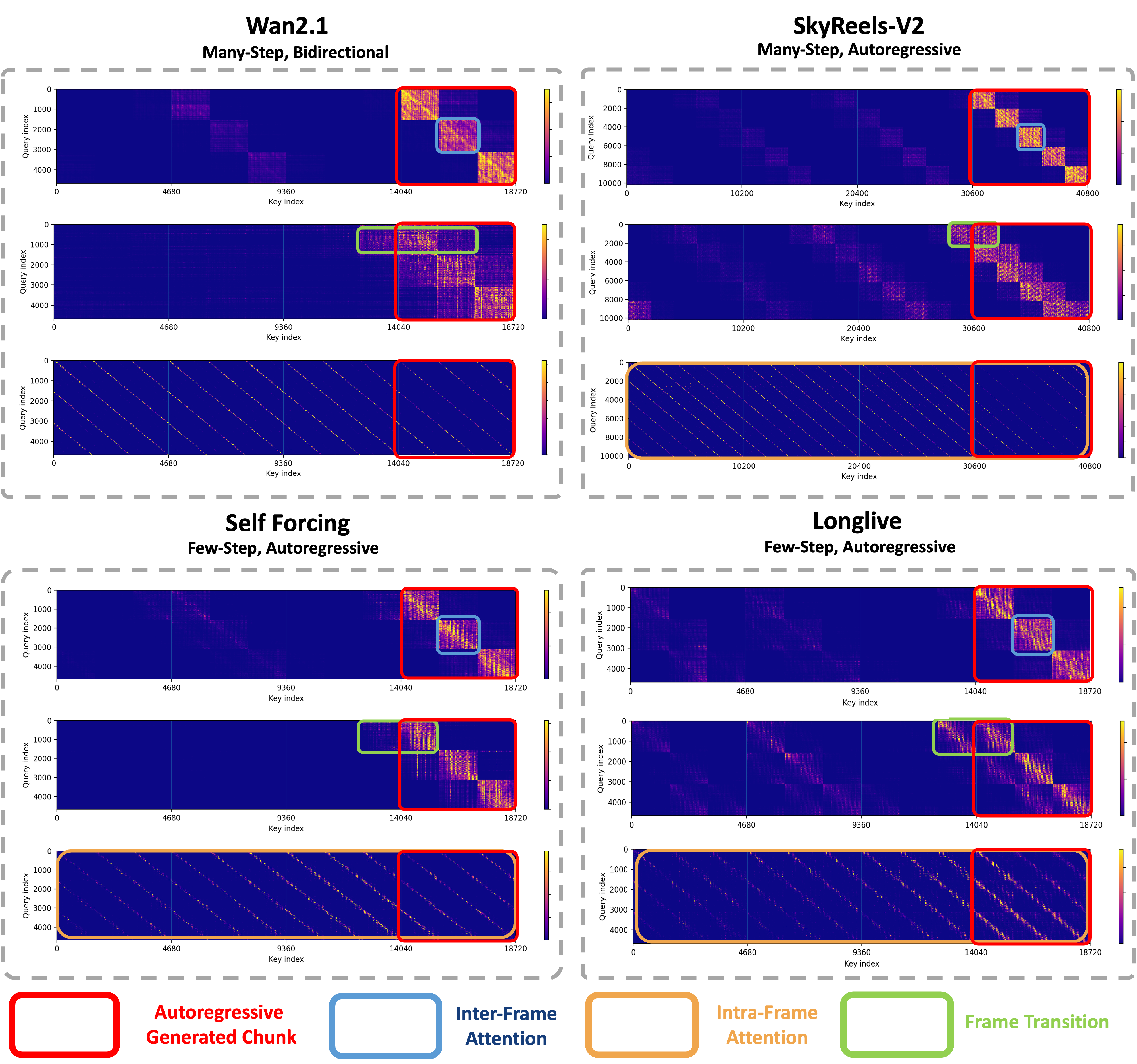}
    \caption{Attention patterns of Wan2.1~\cite{wan2025wan}, SkyReels-V2~\cite{chen2025skyreels}, Longlive~\cite{yang2025longlive}, and Self Forcing~\cite{huang2025selfforcing}.
    % \textbf{Shared Feature:} Head functioning is a shared feature across video generation models of different steps and structures, showing specialization of inter-frame and intra-frame attention. 
    % \textbf{Distinctive Features:} Autoregressive diffusion models exhibit unique characteristics: (i) a strong focus on local frame transitions, and (ii) dependence on variable-length key-value context at each autoregressive step, which creates opportunities for compression.
    }
    \label{fig:attention_map}
\end{figure}

We conduct experiments across bidirectional and autoregressive video diffusion models, including both many-step and few-step variants, such as Wan2.1~\cite{wan2025wan}, SkyReels-V2~\cite{chen2025skyreels}, LongLive~\cite{yang2025longlive}, and Self Forcing~\cite{huang2025selfforcing}, as shown in~\Cref{fig:attention_map}. We find that this spatiotemporal functional specialization of attention heads is a common property across these models, where different heads are respectively responsible for inter-frame and intra-frame attention. Given that mainstream autoregressive video models are typically derived from bidirectional teacher models~\cite{huang2025selfforcing,chen2025skyreels,teng2025magi}, we hypothesize that this property is inherited from the teacher models.

However, autoregressive video models exhibit several distinctive features. First, they learn smooth transitions from the preceding frame. Due to their autoregressive generation process (which follows a Markovian chain), the current block must acquire transition information from previous frames. Our visualization shows that such transition information is primarily concentrated in the transition anchor frame rather than the full history frames. By contrast, in bidirectional models such as Wan, these transitions are typically modeled through bidirectional attention over a local temporal range. 
Second, at each generation step, autoregressive video models condition on previously generated frames, which may contain substantial redundancy. Unlike bidirectional models, autoregressive models do not generate all frames in a single pass and do not require the query and key states to have identical lengths, making KV cache compression over these redundant historical frames a feasible design choice.

\section{User Study}
\label{app:user_study}

\paragraph{Setup.} 
To verify whether the quantitative benchmark results align with human perception, we conducted a user study with 12 participants. Each participant was presented with 15 video groups, where each group contained videos generated by different methods, including the base models (Self Forcing~\cite{huang2025selfforcing} and LongLive~\cite{yang2025longlive}), Dummy Forcing~\cite{guo2026efficient}, and \method (ours). 
The videos covered 5-second and 30-second generations from Self Forcing and LongLive, and were randomly sampled from the full set of 946 videos in the VBench~\cite{huang2023vbench,huang2025vbench++} benchmark.
To avoid positional bias, the videos were randomly arranged as \textit{left}, \textit{middle}, and \textit{right}. In total, we collected 540 evaluations (12 participants × 15 video groups × 3 videos).

\paragraph{Evaluation protocol.}
Following the previous evaluation protocol~\cite{lu2025reward}, we ask the participants to evaluate each video according to three key criteria:
\begin{itemize}[leftmargin=16pt, itemsep=2pt, topsep=2pt]
    \item \textbf{Visual Quality.}
    This criterion captures the overall quality and appeal of each video by jointly considering factors such as visual fidelity, coherence, motion quality, and the subjective viewing experience.

    \item \textbf{Dynamic Degree.}
    This criterion measures the naturalness, richness, and engagement of motions and changes in the video. Participants assess whether the generated content exhibits realistic and diverse dynamics, rather than static or repetitive patterns.

    \item \textbf{Consistency.}
    This criterion assesses whether a video maintains visual quality and coherence throughout its entire duration, without exhibiting visual drift, artifacts, or inconsistencies.
\end{itemize}

\begin{figure}[H]
    \centering
    \includegraphics[width=\linewidth]{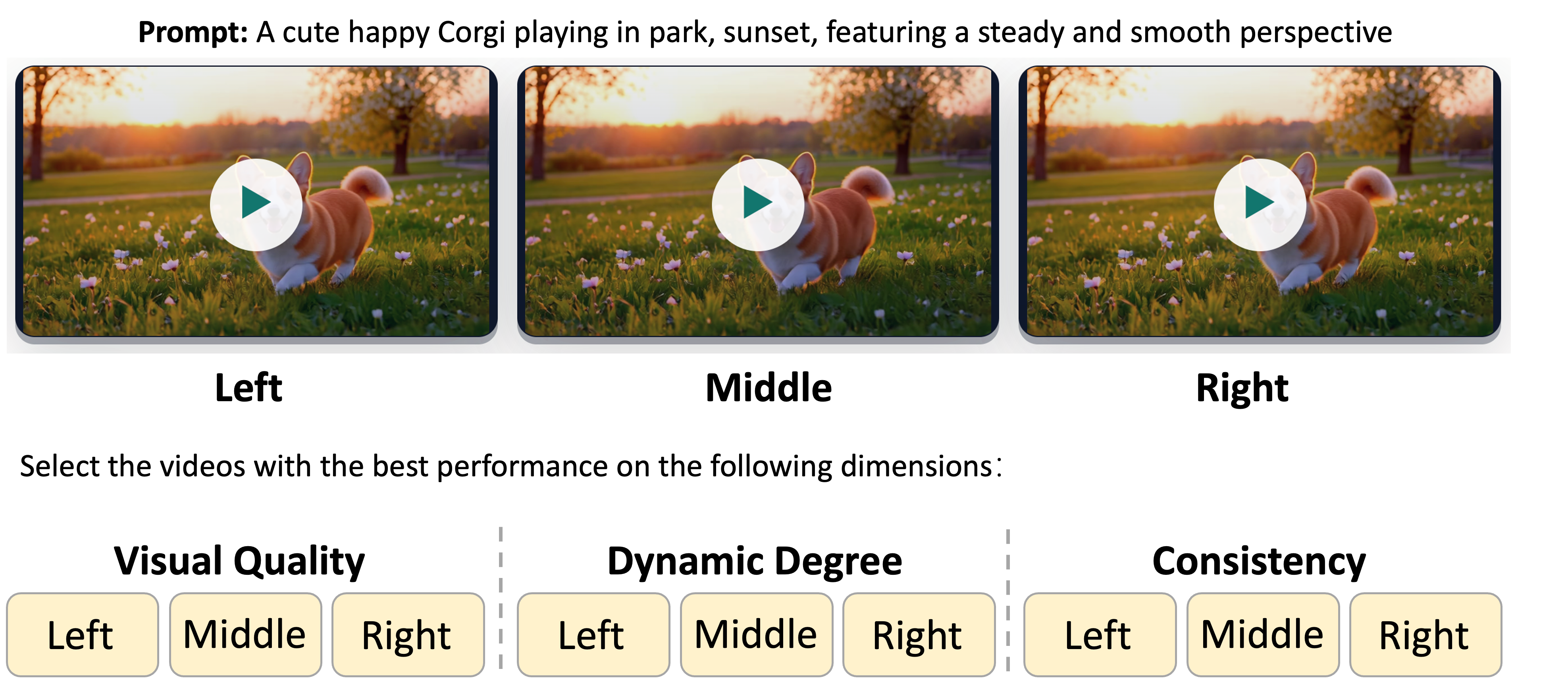}
    \caption{Protocol and screenshot of the user study.
    }
    \label{fig:user_study_protocol}
\end{figure}

\section{Results on Krea-Realtime-14B}
\label{app:krea_14B}

We further extend our method to a larger model scale to validate its effectiveness. Specifically, we evaluate on Krea-Realtime-14B~\cite{erwann2025krea}, a 14B model trained with the \textit{Self Forcing} paradigm. We use only the offline inference version without introducing any additional optimizations. The results in~\Cref{tab:krea_14b} demonstrate that our method remains effective on larger-scale models.

\section{Interactive Video Generation}
\label{app:interactive_result}

We further extend our method to interactive video generation by applying KV cache compression separately to the video segment associated with each prompt, and summarize the resulting quality and efficiency in~\Cref{tab:interactive_result}. Specifically, we use the interactive prompts provided by LongLive~\cite{yang2025longlive}, where each prompt group contains 6 progressively evolving descriptions, each corresponding to a 10-second video segment. We evaluate the quality of the entire generated video using VBench-Long~\cite{huang2025vbench++}. The results show that our method achieves improvements in both quality (79.52 vs. 78.63) and inference speed (26.35 fps vs. 23.07 fps).

\begin{table*}[t]
\centering
\renewcommand{\arraystretch}{1.12}
\setlength{\tabcolsep}{4.5pt}

\resizebox{\linewidth}{!}{%
\begin{tabular}{lcc|cc|ccc}
\toprule
\multirow{2}{*}{Method}
& \multicolumn{2}{c|}{Efficiency Metrics}
& \multicolumn{2}{c|}{Core Metrics}
& \multicolumn{3}{c}{General Metrics} \\
\cmidrule(lr){2-3}
\cmidrule(lr){4-5}
\cmidrule(lr){6-8}
& FPS$\uparrow$
& Speedup$\uparrow$
& \makecell{Chunk\\Disc.$\downarrow$}
& \makecell{Dynamic\\Degree$\uparrow$}
& \makecell{Quality\\Score$\uparrow$}
& \makecell{Semantic\\Score$\uparrow$}
& \makecell{Total\\Score$\uparrow$} \\
\midrule

{\color{gray}Full KV}
& {\color{gray}4.13} & {\color{gray}1.00$\times$}
& {\color{gray}1.7} & {\color{gray}77.78}
& {\color{gray}85.02} & {\color{gray}81.57} & {\color{gray}84.33} \\

\rowcolor{ORANGEII!20}
\textbf{\method\ (Ours)}
& 5.22 & \textbf{1.26$\times$} & 1.9 & 73.61 & 85.35 & 81.55 & \textbf{84.59} \\

% 5.18 68.06
%   "total score": 0.8503742317434433,
%   "semantic_score": 0.8159781440293804,
%   "total_score": 0.8434950142006308

\bottomrule
\end{tabular}%
}

\caption{VBench results on 5-second video generation with Krea-Realtime-14B.}
\label{tab:krea_14b}
\end{table*}

\begin{table*}[t]
\centering
\renewcommand{\arraystretch}{1.3}

\resizebox{\textwidth}{!}{%
\begin{tabular}{l|c|ccccccc}
\toprule
\multirow{2}{*}{Methods}
& \multirow{2}{*}{FPS$\uparrow$}
& \multicolumn{7}{c}{Quality Score} \\
\cmidrule(lr){3-9}
&
& \makecell{Total$\uparrow$}
& \makecell{Imaging\\Quality$\uparrow$}
& \makecell{Subject\\Consistency$\uparrow$}
& \makecell{Background\\Consistency$\uparrow$}
& \makecell{Motion\\Smoothness$\uparrow$}
& \makecell{Dynamic\\Degree$\uparrow$}
& \makecell{Aesthetic\\Quality$\uparrow$} \\
\midrule

Longlive
& 23.07
& 78.63 & 69.38 & 98.03 & 96.21 & 99.14 & 26.39 & 59.42 \\

Dummy Forcing
& 25.74
& 78.04 & 69.42 & 96.30 & 95.01 & 98.57 & 30.28 & 59.80 \\
    
\rowcolor{ORANGEII!20}
\textbf{\method}
& \textbf{26.35}
& \textbf{79.52} & 70.13 & 97.51 & 95.89 & 98.89 & 36.11 & 60.58 \\

\bottomrule
\end{tabular}%
}

\caption{VBench-Long results on 60-second interactive video generation with Longlive.}
\label{tab:interactive_result}
\end{table*}

\begin{table}[H]
\captionsetup{justification=centering,skip=6pt}
\centering
\renewcommand{\arraystretch}{1.2}
\setlength{\tabcolsep}{4pt}

\resizebox{0.7\linewidth}{!}{%
\begin{tabular}{lcc|ccc}
\toprule
\multirow{2}{*}{Methods}
& \multicolumn{2}{c|}{Efficiency Metrics}
& \multicolumn{3}{c}{Performance Metrics} \\
\cmidrule(lr){2-3}
\cmidrule(lr){4-6}
& FPS$\uparrow$
& Speedup$\uparrow$
& \makecell{Chunk\\Disc.$\downarrow$}
& \makecell{Dynamic\\Degree$\uparrow$}
& \makecell{Total\\Score$\uparrow$} \\
\midrule

\multicolumn{6}{c}{\textbf{LongLive - 60s}} \\
\midrule

\textbf{\method}
& 26.71 & 1.30$\times$
& 2.5 & 43.56 & 80.43 \\

\textbf{\method\ + FP8}
& 28.05 & 1.37$\times$
& 2.5 & 43.27 & 80.41 \\

\midrule
\multicolumn{6}{c}{\textbf{Self Forcing - 30s}} \\
\midrule

\textbf{\method}
& 26.65 & 1.50$\times$
& 2.7 & 52.23 & 80.07 \\

\textbf{\method + FP8}
& 27.55 & 1.55$\times$
& 2.8 & 52.15 & 80.06 \\

\bottomrule
\end{tabular}%
}
\caption{Quality and efficiency results with FP8 quantization.}
\label{tab:fp8}
\end{table}

\section{Seamless Integration of Quantization}
\label{app:quantization}

We further incorporate FP8 quantization~\cite{zhang2024sageattention2}
into our KV cache compression method. This optimization reduces the computational cost of attention module by leveraging FP8 attention kernels tailored to the NVIDIA Hopper architecture, which further boosts throughput with minimal performance drop, as shown in~\Cref{tab:fp8}. This shows that our method is compatible with other acceleration techniques.

\section{More Ablation Study Results}
\label{app:more_ablation}
In~\Cref{sec:ablation_study}, we present the ablation results of LongLive on 30-second videos. For completeness, \Cref{tab:ablation_longlive_5s,tab:ablation_selfforcing_5s} further report the ablation results on Self Forcing and LongLive for 5-second videos. The overall trends \textbf{remain consistent}. 
Regarding head profiling, our simple profiling strategy is already sufficient to closely approach manual profiling, while significantly outperforming the unprofiled random baseline. 
For KV cache compression, removing the transition anchor frame (\textit{w/o static cache}) leads to severe chunk discontinuity, which further degrades the overall score. 
In contrast, removing the KV cache for dynamic heads (\textit{w/o dynamic cache}) mainly causes a loss in dynamics, while having little effect on chunk continuity and the overall score. 
Since 5-second videos do not accumulate sufficient context, these effects are less pronounced than those observed on 30-second videos.

\begin{table*}[t]
\captionsetup{justification=raggedright,singlelinecheck=false}
\centering
\renewcommand{\arraystretch}{1.15}
\setlength{\tabcolsep}{5pt}

\begin{minipage}[t]{0.48\textwidth}
\captionsetup{justification=raggedright,singlelinecheck=false}
\centering
\resizebox{\textwidth}{!}{%
\begin{tabular}{lccc}
\toprule
Method
& \makecell{Chunk\\Disc.$\downarrow$}
& \makecell{Dynamic\\Degree$\uparrow$}
& \makecell{Total\\Score$\uparrow$} \\
\midrule

\multicolumn{4}{c}{\textit{Head Profiling}} \\
\midrule

\textbf{\method}
& 2.1 & 45.56 & 83.22 \\

Random Profiling
& 2.5 & 41.94 & 82.82 \\

Human Profiling
& 2.2 & 47.22 & 83.31 \\
\midrule

\multicolumn{4}{c}{\textit{KV Cache Compression}} \\
\midrule

\textbf{\method}
& 2.1 & 45.56 & 83.22 \\

w/o static-head cache
& 3.9 & 42.22 & 82.60 \\

w/o dynamic-head cache
& 2.3 & 43.33 & 83.08 \\

\bottomrule
\end{tabular}%
}
\caption{Ablation study on KV cache compression and head profiling strategies on LongLive - 5s.}
\label{tab:ablation_longlive_5s}
\end{minipage}
\hfill
\begin{minipage}[t]{0.48\textwidth}
\captionsetup{justification=raggedright,singlelinecheck=false}
\centering
\resizebox{\textwidth}{!}{%
\begin{tabular}{lccc}
\toprule
Method
& \makecell{Chunk\\Disc.$\downarrow$}
& \makecell{Dynamic\\Degree$\uparrow$}
& \makecell{Total\\Score$\uparrow$} \\
\midrule

\multicolumn{4}{c}{\textit{Head Profiling}} \\
\midrule

\textbf{\method}
& 2.1 & 69.17 & 83.98 \\

Random Profiling
& 3.0 & 59.72 & 82.76 \\

Human Profiling
& 2.2 & 71.39 & 84.10 \\

\midrule

\multicolumn{4}{c}{\textit{KV Cache Compression}} \\
\midrule

\textbf{\method}
& 2.1 & 69.17 & 83.98 \\

w/o static-head cache 
& 3.4 & 65.28 & 83.73 \\

w/o dynamic-head cache
& 2.8 & 62.50 & 83.56 \\

\bottomrule
\end{tabular}%
}
\caption{Ablation study on KV cache compression and head profiling strategies on Self Forcing - 5s.}
\label{tab:ablation_selfforcing_5s}
\end{minipage}
\end{table*}

\section{Implementation Details}
\label{app:implementation_detail}

\paragraph{Baselines.}
We mainly compare our method with KV cache compression approaches for large language models, StreamingLLM~\cite{xiaoefficient}, and with Dummy Forcing~\cite{guo2026efficient}, a representative KV cache compression method for AR diffusion models.
For StreamingLLM, we implement a naive frame-level strategy that retains 3 sink frames and 4 recent frames in the historical KV cache. For Dummy Forcing, following the original paper, we preserve the KV cache of the local region for local heads and adjust the history cache length $L$ for neighbor heads. For an aggressive variant, we follow the official repository and set $L=1$. For a conservative variant, we retain $L=6$ for Self Forcing~\cite{huang2025selfforcing} and $L=2$ for LongLive~\cite{yang2025longlive}, corresponding to their original window sizes.
A larger value of $L$ provides a broader visible context, which can improve performance at the cost of reduced speed.

\paragraph{Our Method.}
Our method consists of two components, head profiling and hybrid KV cache compression, corresponding to the hyperparameter $\alpha$ and the compression ratio $r$ for dynamic heads, respectively. For head profiling, we perform offline head classification using a single prompt with $\alpha=0.8$, and the entire procedure can be completed within a few minutes. 
For hybrid KV cache compression, we retain transition frames for static heads, while setting the default compression ratio for dynamic heads to $r=0.3$. Specifically, each historical frame in the KV cache is uniformly divided into $n=6$ contiguous segments, segment-wise similarity is computed between adjacent frames, and the retained segments are then determined according to the compression ratio $r$.
To avoid the lack of adjacent frames at the beginning of generation, we apply KV cache compression starting from the second autoregressive step. We preserve the sink frame similar to Dummy Forcing~\cite{guo2026efficient}.

\section{VBench Scores Across All Dimensions}
\label{app:vbench_all_dimension}

We further provide the detailed 16-dimension VBench~\cite{huang2023vbench} scores of short videos in~\Cref{tab:main_result_short}.
Specifically, on individual metrics, \method achieves improved performance over its base model in dynamic degree, overall consistency, appearance style, and total score with reduced KV cache budget.
We attribute this result primarily to the fact that \method preserves the context required by each type of head during inference, without sacrificing either local transition information or historical contextual information. 
Another possible factor is that existing autoregressive models may still be imperfect in their ability to utilize distant context effectively.

\begin{figure}[H]
    \centering
    \includegraphics[width=1.0\linewidth]{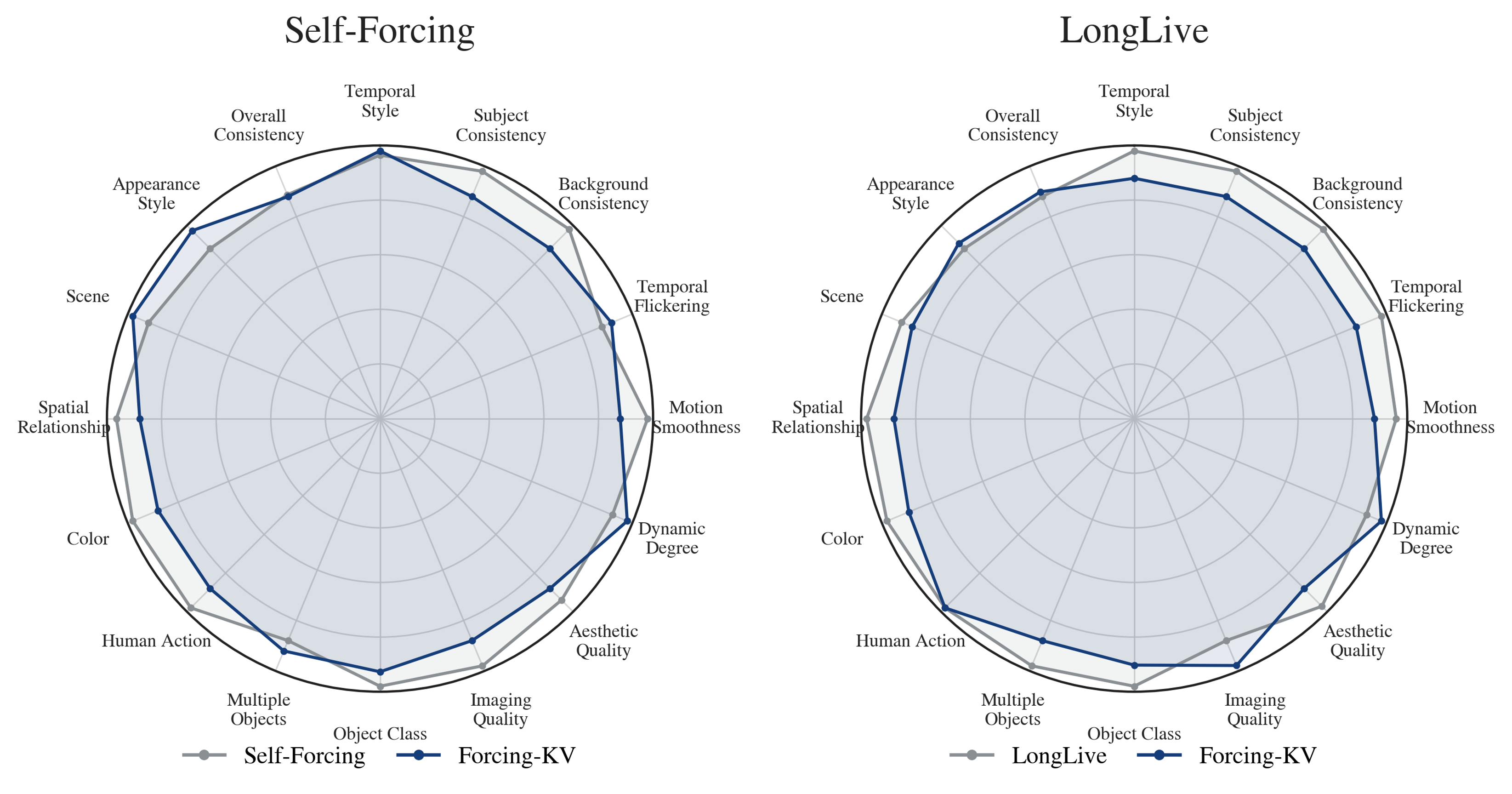}
    \caption{Visualization of VBench~\cite{huang2023vbench} scores. We compare \method with its base model.
    \method achieves near-lossless performance in terms of total score, and further outperforms the baseline on metrics such as dynamic degree, overall consistency, and appearance style, demonstrating its advantage.
    }
    \label{fig:raider}
\end{figure}

\section{Where Are Static and Dynamic Heads Located?}
\label{app:head_distribution}
\begin{figure}[H]
    \centering
    \includegraphics[width=\linewidth]{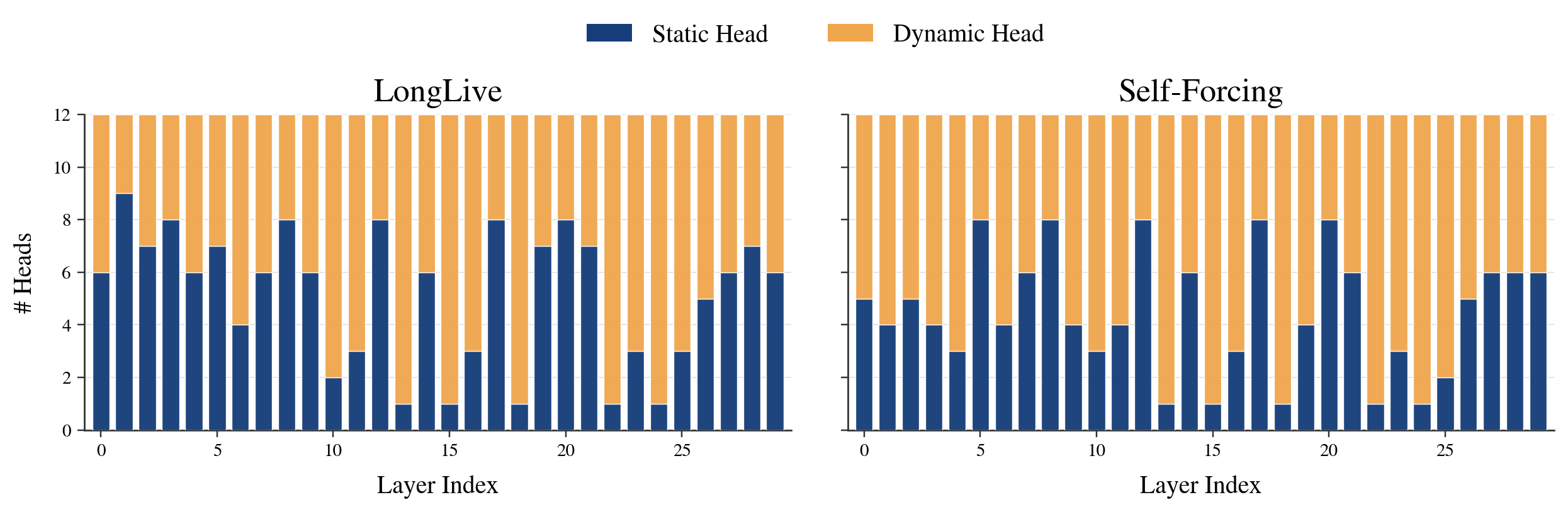}
    \caption{Head distribution across layers.}
    \label{fig:head_distribution}
\end{figure}

We consider the head distribution in AR diffusion models to be an interesting research problem, and visualize the distribution of heads across layers in~\Cref{fig:head_distribution}. 
Overall, dynamic heads constitute approximately 60\% of all heads, with a noticeably higher proportion in the middle layers (layer indices 13, 15, and 17). 
We hypothesize that this pattern arises because layers near the input and output focus more on extracting structured information to preserve local video quality, whereas the intermediate layers make heavier use of contextual information for rich feature refinement, improving detail consistency and temporal dynamics.

\section{Trend of the Proportion of Self-Attention}
In~\Cref{fig:scaling_law}, we show how the total runtime increases as the attention window size and resolution grow. As a complementary analysis, we further break down the proportion of time consumed by self-attention. As the sequence length grows, self-attention gradually occupies a larger share of the overall Transformer block (24\% to 61\% and 61\% to 89\%), making KV cache compression increasingly beneficial.

\begin{figure}[H]
    \centering
    \includegraphics[width=\linewidth]{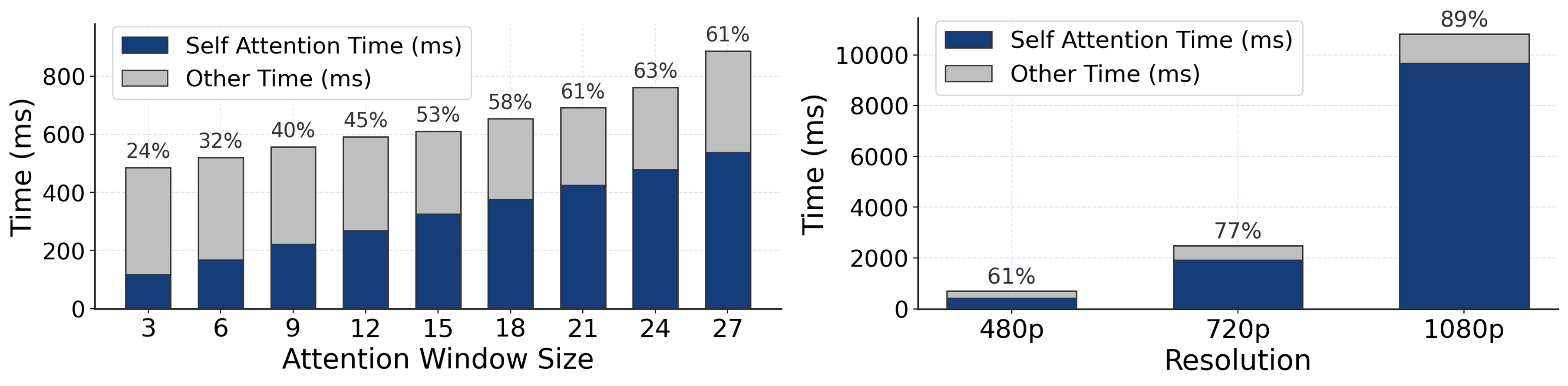}
    \caption{Variation in the proportion of total runtime occupied by self-attention.}
    \label{fig:attention_portion}
\end{figure}

\section{Quality Examples}
\label{app:quality example}

\begin{figure}[H]
    \centering
    \includegraphics[width=\linewidth]{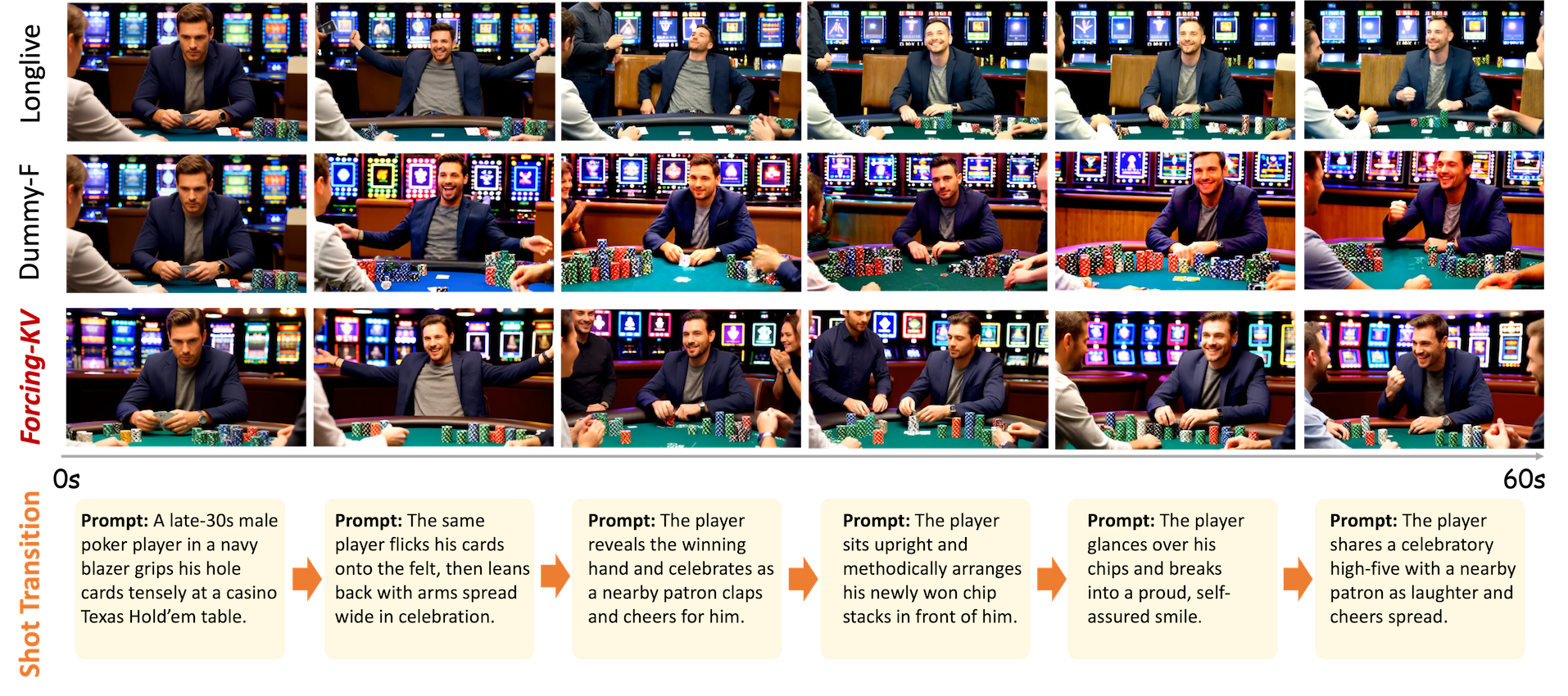}
    
    \vspace{20pt}
    
    \includegraphics[width=\linewidth]{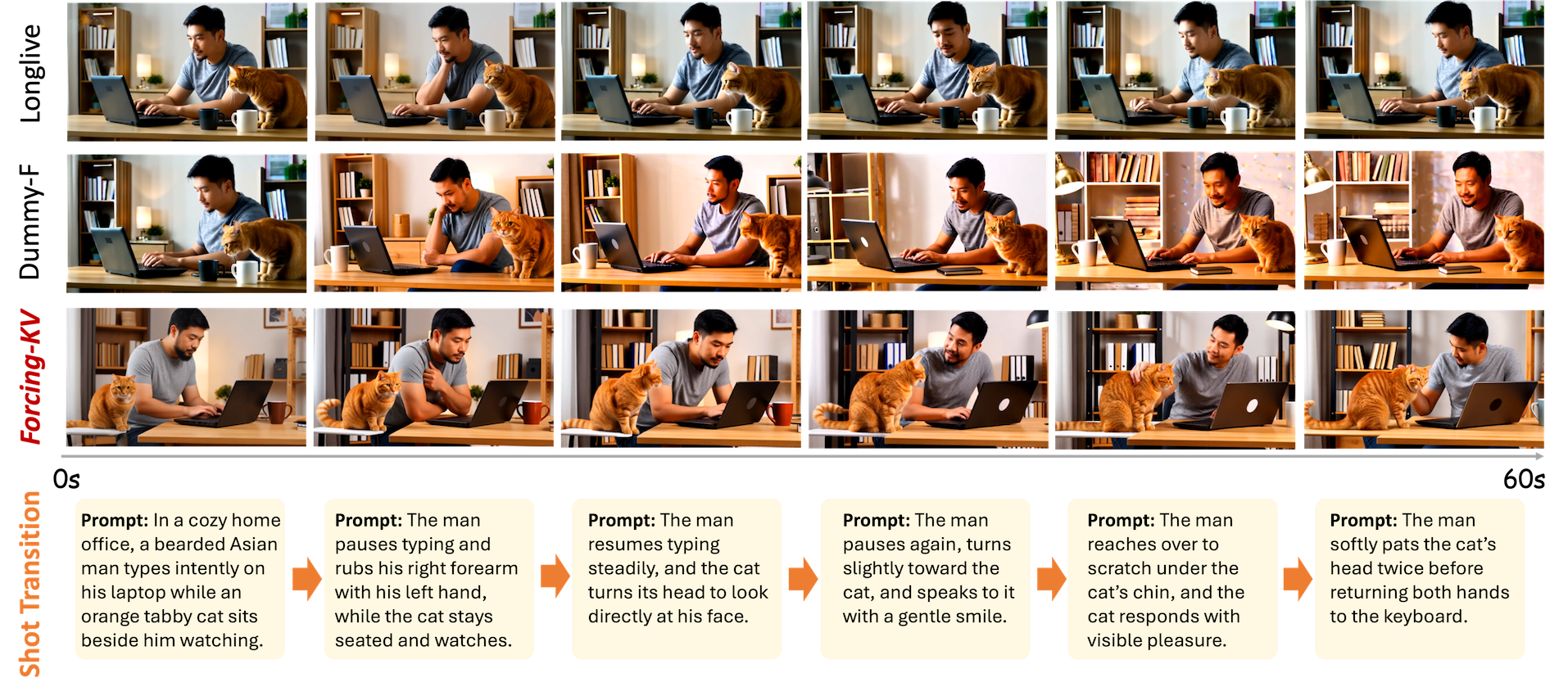}
    \caption{Quality example of 60-second interactive video on Longlive.}
    \label{fig:case_inter}
\end{figure}

\clearpage

\begin{figure}[H]
    \centering
    \includegraphics[width=0.87\linewidth]{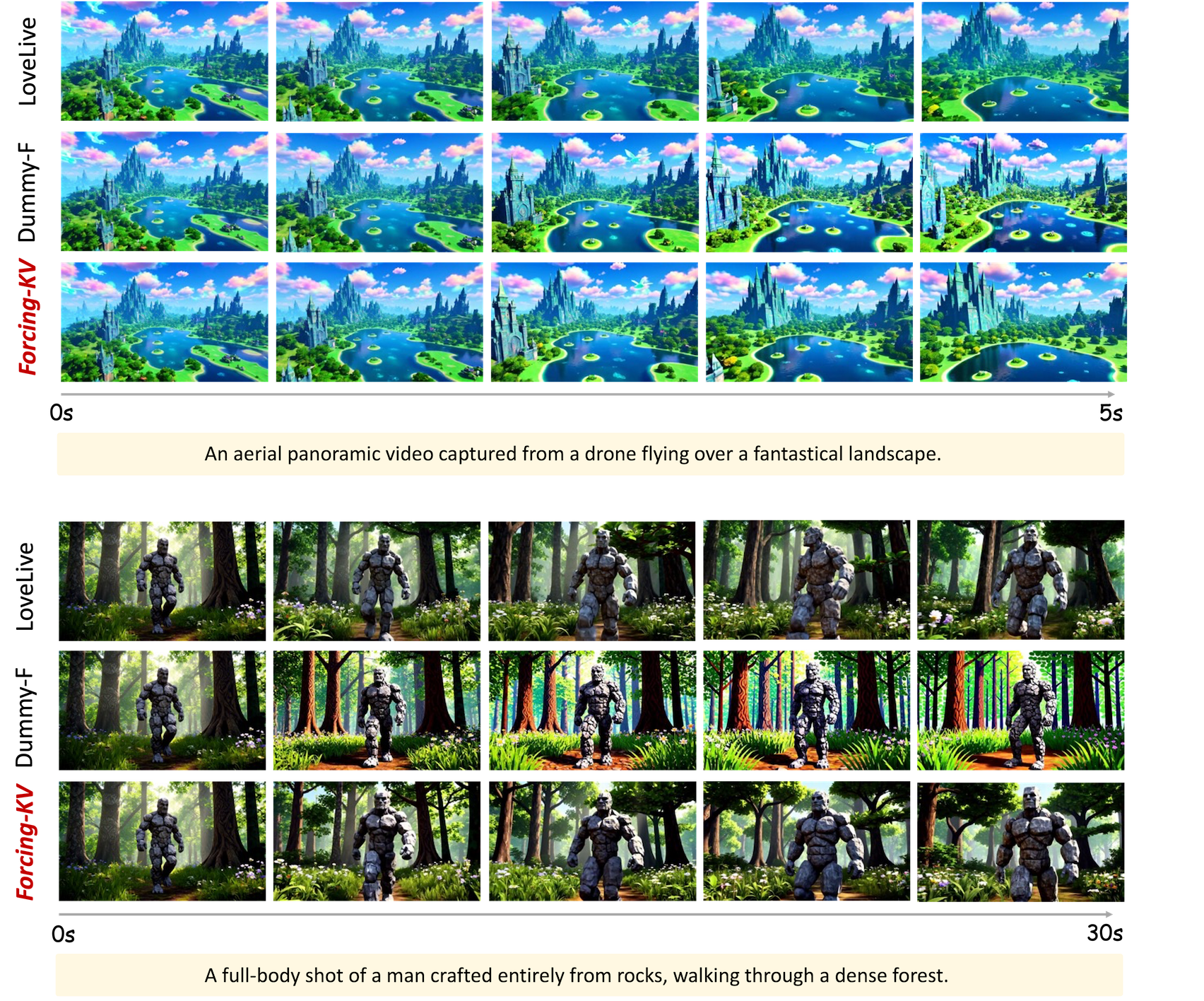}
    \caption{Quality examples on Longlive 5s and 30s.}
    \label{fig:case_study1}
\end{figure}

\begin{figure}[H]
    \centering
    \includegraphics[width=0.85\linewidth]{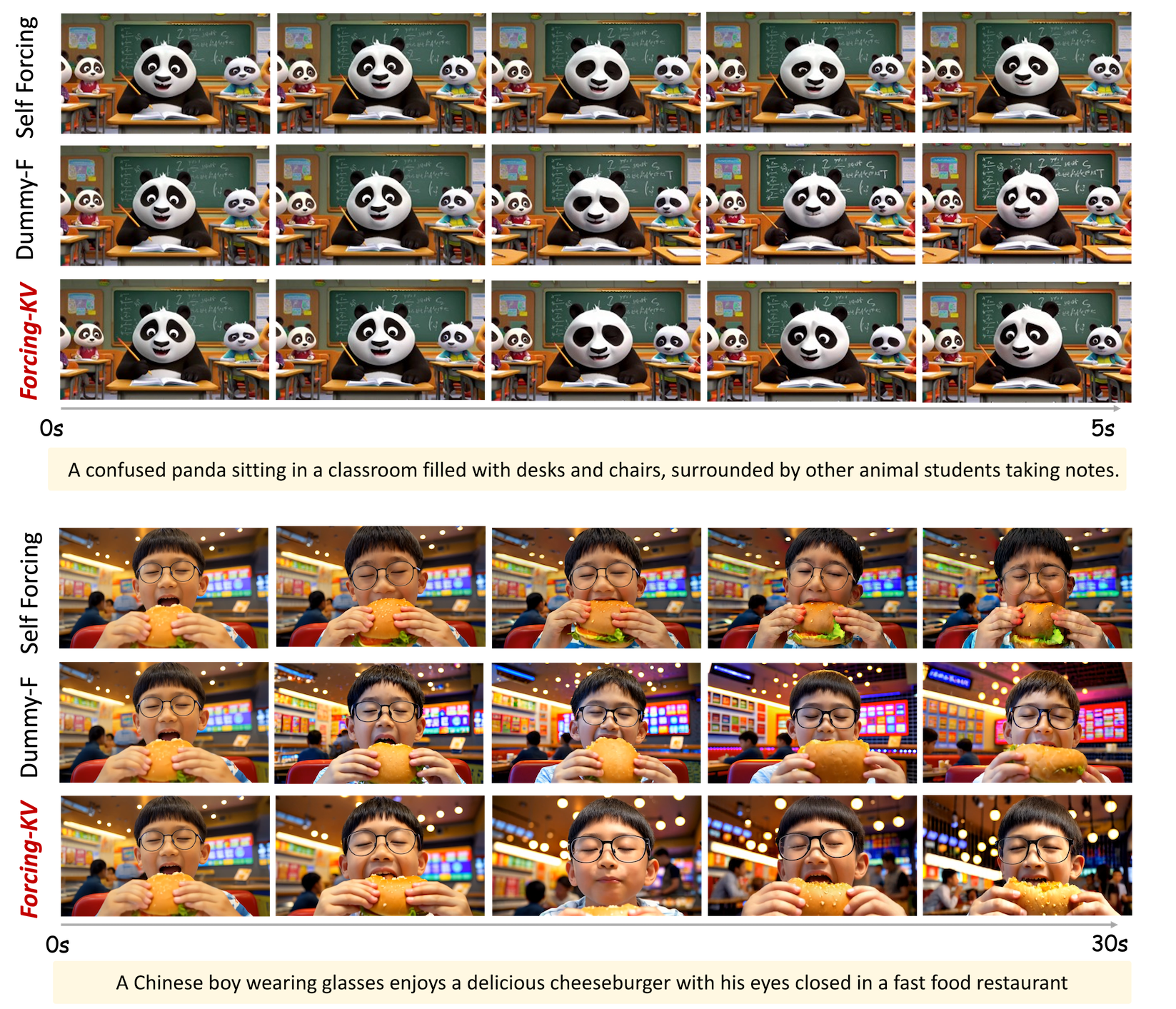}
    \caption{Quality examples on Self Forcing 5s and 30s.}
    \label{fig:case_study2}
\end{figure}

% \newpage
% \input{checklist.tex}

\end{document}